\documentclass[journal]{IEEEtran}
\usepackage{amsmath}
\usepackage{amssymb}
\usepackage{amsthm}
\usepackage{array}
\usepackage{amsfonts}
\usepackage{multirow}
\usepackage{xcolor,colortbl}
\usepackage{hhline}
\usepackage{booktabs}
\usepackage{url}
\usepackage{color}
\usepackage{float}
\usepackage{graphicx}
\usepackage{caption}
\usepackage{pgfplots}
\usepackage{subcaption}
\usepackage{algorithm}
\usepackage{stfloats}
\usepackage{multicol}
\usepackage{algorithmicx, tabularx}
\usepackage{algpseudocode}
\usepackage{threeparttable}
\usepackage{hyperref}
\usepackage{lettrine}
\usepackage{xcolor}
\usepackage{soul}
\usepackage{cleveref}
\usepackage[numbers,sort&compress]{natbib}
\usepackage[font=small]{caption}

\algnewcommand\algorithmicinput{\textbf{Input:}}
\algnewcommand\algorithmicoutput{\textbf{Output:}}
\algnewcommand\INPUT{\item[\algorithmicinput]}
\algnewcommand\OUTPUT{\item[\algorithmicoutput]}
\algnewcommand\algorithmicinit{\textbf{Initialize:}}
\algnewcommand\Init{\item[\algorithmicinit]}

\newcommand{\norm}[1]{\left\lVert#1\right\rVert}

\DeclareMathOperator*{\argmax}{arg\,max}

\algnewcommand\algorithmicforeach{\textbf{for each}}
\algdef{S}[FOR]{ForEach}[1]{\algorithmicforeach\ #1\ \algorithmicdo}

\newcommand{\etal}{\textit{et al. }}

\makeatletter
\newcommand{\multiline}[1]{%
  \begin{tabularx}{\dimexpr\linewidth-\ALG@thistlm}[t]{@{}X@{}}
    #1
  \end{tabularx}
}
\makeatother

\hypersetup{
    colorlinks=true,
    linkcolor=blue,
    anchorcolor = blue,
    citecolor = blue,
    filecolor=blue,      
    urlcolor=blue,
    pdftitle={Sharelatex Example},
    pdfpagemode=FullScreen,
    }
    
\begin{document}
\raggedbottom
\title{Multitask Neuroevolution for Reinforcement Learning with Long and Short Episodes}

\author{Nick Zhang, Abhishek Gupta, Zefeng Chen, and Yew-Soon Ong

\thanks{Nick Zhang is with the School of Computer Science and Engineering, Nanyang Technological University, Singapore 639798. (Email: nick0007@e.ntu.edu.sg)}
\thanks{Abhishek Gupta is with the Singapore Institute of Manufacturing Technology (SIMTech), Agency for Science, Technology and Research (A*STAR), Singapore 138634. He also holds a secondary appointment with the School of Computer Science and Engineering, Nanyang Technological University. (Email: abhishek\_gupta@simtech.a-star.edu.sg, abhishekg@ntu.edu.sg.)}
\thanks{Zefeng Chen is with the School of Artificial Intelligence, Sun Yat-sen University, China, and also with the School of Computer Science and Engineering, NTU, Singapore (e-mail: chenzef5@mail.sysu.edu.cn; zefeng.chen@ntu.edu.sg).}
\thanks{Yew-Soon Ong is with the Data Science and Artificial Intelligence Research Centre, School of Computer Science and Engineering, NTU, Singapore, and also the A*Star Center for Frontier AI Research, (Email: asysong@ntu.edu.sg, Ong\_Yew\_Soon@hq.a-star.edu.sg)} }

\maketitle

\begin{abstract}
Studies have shown evolution strategies (ES) to be a promising approach for reinforcement learning (RL) with deep neural networks. However, the issue of high sample complexity persists in applications of ES to deep RL over long horizons. This paper is the first to address the shortcoming of today's methods via a novel \emph{neuroevolutionary multitasking} (NuEMT) algorithm, designed to transfer information from a set of auxiliary tasks (of short episode length) to the target (full length) RL task at hand. The auxiliary tasks, extracted from the target, allow an agent to update and quickly evaluate policies on shorter time horizons. The evolved skills are then transferred to guide the longer and harder task towards an optimal policy. We demonstrate that the NuEMT algorithm achieves data-efficient evolutionary RL, reducing expensive agent-environment interaction data requirements. Our key algorithmic contribution in this setting is to introduce, for the first time, a multitask skills transfer mechanism based on the statistical \emph{importance sampling} technique. In addition, an adaptive resource allocation strategy is utilized to assign computational resources to auxiliary tasks based on their gleaned usefulness. Experiments on a range of continuous control tasks from the OpenAI Gym confirm that our proposed algorithm is efficient compared to recent ES baselines.

\end{abstract}


\section{Introduction}
In reinforcement learning (RL), agents learn how to interact with dynamic and uncertain environments by taking actions that maximize expected rewards. To this end, many popular RL algorithms, such as TRPO \cite{SchulmanLAJM15}, DDPG \cite{LillicrapHPHETS15}, D4PG \cite{Barth-MaronHBDH18}, DQN \cite{MnihKSRVBGRFOPB15}, among others, are based on the Markov decision process (MDP) formalism and the concept of value functions. Algorithms belonging to this class have achieved notable success in applications such as locomotion tasks \cite{SchulmanLAJM15, SchulmanWDRK17}, strategy board games (e.g., surpassing the level of the world champion in GO \cite{Silver2016}) and playing Atari from pixels \cite{MnihKSRVBGRFOPB15, SchulmanLAJM15, SchulmanWDRK17}.

Advances in RL have however shed light on \emph{evolution strategies} (ES) as a viable alternative for handling continuous control tasks \cite{igel2003neuroevolution}, or even playing Atari games with pixel inputs \cite{SalimansHCS17}. Being a derivative-free approach, ES attempts to directly search for an optimal policy mapping the states of an environment to the best course of action, with the objective of maximizing the agent's cumulative rewards. This type of approach falls under the umbrella of \emph{neuroevolution} when the underlying policy is parameterized by a neural network \cite{RisiT17}. The simplicity of implementation (obviating the need to propagate derivatives through the computation graph), better exploration capacity, and inherently parallelizable nature of population-based search makes ES a scalable option for RL, especially given access to distributed computer systems.

Despite the simplicity, surprising levels of performance achievable, and scalability of ES---delivering linear speedup in the number of CPU cores available---the high sample complexity of such black-box optimization methods leaves much to be desired. Even though ES can make good use of parallel resources, the need for vast amounts of agent-environment interaction data to evaluate populations of candidate policies can become prohibitively resource-intensive. Moreover, each data sample generated in a standard ES is utilized just for a single search update, hence indicating a wastage of costly and useful information. The issue of lowering sample complexity has thus attracted significant research interest lately, with various methods proposed for making effective use of data in the search for optimal policies \cite{liu2019trust, GangwaniP18}. One promising approach in this regard is that of \emph{experience transfer} \cite{Yu18}. The motivation derives from an analogy to humans who, instead of solving problems from scratch, learn to adapt and reuse experiential priors. Likewise, it is deemed that RL agents may also learn more efficiently by building on data/experiences drawn from \emph{related} tasks. Algorithmic realizations of this concept have even been explored beyond the realms of RL, for teaching machines transferable skills in diverse areas of learning \cite{zhuang2020comprehensive} and optimization \cite{GuptaOF18}.

Tapping on experience transfers in deep RL, Fuks \etal \cite{FuksAHL19} proposed the idea of \emph{progressive episode lengths} (PEL) with canonical ES \cite{ChrabaszczLH18}. PEL comprises a set of artificially generated auxiliary tasks, extracted from the target RL environment, that are distinguished by the ascending order of their episode lengths. The tasks are then tackled \emph{sequentially}, with the search for each task seeded by good solutions found for its predecessor. This procedure naturally leads to the reuse of experiential priors from related (short time horizon) tasks, with the hope of speeding up learning on longer and harder tasks. However, the direct seeding of solutions could cause harmful \emph{negative transfers}. Evolved skills may overspecialize to an auxiliary task, and hence not generalize well to the target (full length) task at hand. As illustration, imagine a hypothetical agent to be trained for a 26.2-mile marathon, where energy is to be strategically conserved. If such an agent first specializes on shorter 100-metre sprints, then the burst in speed would be energy sapping, leaving little in the tank for the full marathon.

In this paper, we thus re-examine experience transfers for the first time in the light of \emph{evolutionary multitasking} (EMT) \cite{GuptaOF16, zhang2021study}, where an evolving population is uniquely leveraged to tackle multiple tasks (auxiliary and the target) \emph{concurrently} with periodic exchange of generated solutions. Our aim is to achieve data-efficient evolutionary RL by effectively exploiting transferable skills while guarding against threats of negative transfer. Minimizing wastage of computational resources on auxiliary tasks that are not useful to the target is a key consideration. To this end, we propose a new EMT algorithm that builds on a popular variant of ES (labelled as the OpenAI-ES \cite{SalimansHCS17}) with known efficacy for direct neural network policy search. We refer to our \emph{neuroevolutionary multitasking} algorithm as NuEMT for short. \emph{The crucial distinction from \cite{FuksAHL19} is that instead of processing tasks sequentially, NuEMT combines all auxiliary (short episode length) tasks with the target in a single multitask formulation. This not only supports sample efficiency by learning from relevant experiential priors, but also enables online neutralization of those tasks whose evolved skills do not transfer well to the target agent}. 

Since the inception of EMT, a range of techniques using \emph{probability mixture models} (to capture distributional overlaps of jointly evolving populations) have been proposed to adapt the quantity and frequency of inter-task information transfers \cite{BaliOGT20}, with successful application in \cite{Nick2021, bali2020cognizant, shang2022solving}. However, the need to rebuild mixture models at prespecified ``transfer intervals" can be expensive, bringing an additional layer of internal algorithmic complexity to the search. To overcome this bottleneck, we equip NuEMT with a novel yet simple multitask transfer mechanism based on \emph{importance sampling} \cite{tokdar2010importance}. Stochastic gradient estimates with cross-task solution sampling and fast mixture model updates is made possible, without the need to continually rebuild the model from the ground up. What's more, stochastic updates of the mixture coefficients yield the evolving relevance of the auxiliary tasks. Hence, the coefficient values are used to allocate more computational resources to those tasks that are gleaned to be useful to the target, while neutralizing those that are not.

In sum, the main contributions of this paper, including the problem setup and the proposed algorithm, are fourfold:
\begin{itemize}
  \item An EMT formulation with skills transfer from tasks with shorter to longer episode lengths is presented.
  \item A novel NuEMT algorithm with importance sampling for inter-task experience transfer is crafted.
  \item An online resource allocation strategy is  embedded in NuEMT to minimize wastage of computational resources on those tasks that are unaligned with the target. 
  \item A series of experimental studies confirm the effectiveness of NuEMT in reducing agent-environment interaction data requirements in evolutionary RL.
\end{itemize}

The remainder of this paper is organized as follows. Section II contains a brief overview of related work in the literature. Section III contains the preliminaries of ES, EMT, and importance sampling. Section IV describes the generation of auxiliary tasks in RL and presents the NuEMT algorithm. In Section V, we test the algorithm and compare its performance against recent ES baselines. We conclude the paper with directions for future research in Section VI. 

\section{Related Work}
There is growing interest in neuroevolutionary algorithms for direct policy search in RL. One of the most notable works in this regard is that by Salimans \etal \cite{SalimansHCS17}, which established ES as a viable alternative for deep RL. Their experiments revealed the advantages of ES in terms of better exploration capacity (compared to a policy gradient method TRPO \cite{SchulmanLAJM15}) and the ease of scaling to thousands of parallel workers. (This variant of ES has since come to be known as OpenAI-ES in the literature \cite{PagliucaMN20, ChenZHJ19}, and will be referred as such hereinafter.) However, even though the runtime of a population-based ES can be greatly reduced by distributing workloads on modern distributed computer systems, the high sample complexity (requiring vast amounts of environment interaction data for policy updates) remains a computational bottleneck.

Earlier works in ES also evinced its potential applicability for direct policy search. In particular, Igel \cite{igel2003neuroevolution} successfully demonstrated the application of the CMA-ES \cite{hansen2003reducing} for RL. Two variable metric methods for solving RL tasks, namely, the \emph{natural actor critic} algorithm \cite{PetersS07} and the CMA-ES, were compared and contrasted in \cite{Heidrich-Meisner_Igel08}. The results showed CMA-ES to be more robust to the selection of hyper-parameters, while being competitive in terms of learning speed.

More recently, significant research effort has been aimed at lowering the sample complexity of ES by introducing better search directions, effective utilization of historical data, and exploration techniques such as \emph{novelty search} \cite{gomes2015devising}. Choromanski \etal \cite{ChoromanskiRSTW18} showed  that random orthogonal and Quasi Monte Carlo finite difference directions could be more effective for parameter exploration than the random Gaussian directions in \cite{SalimansHCS17}. Liu \etal \cite{LiuLQ20} proposed to improve sample efficiency by reducing the variance of the stochastic gradient estimator of the vanilla ES in high-dimensional optimization. This was done by sampling search directions from a hybrid probabilistic distribution characterized by a gradient subspace---defined by recent historical estimated gradients---and its orthogonal complement. In \cite{liu2019trust}, efficient use of sampled data was sought by an iterative procedure that optimizes a surrogate objective function. A monotonic improvement guarantee for such procedure was theoretically proven. Further, Conti \etal \cite{ContiMSLSC18} hybridized novelty search with ES to enhance policy space exploration, encouraging RL agents to exhibit different behaviors that reduce the danger of being stuck indefinitely (hence wastefully) in local optima of deceptive reward functions.

Neuroevolution with genetic algorithms (GAs)---that differ from ES mainly in that they employ crossover operators---has also shown promising results in tuning the parameters of deep neural network policies. Such \etal \cite{Such17} investigated the use of a simple GA on deep RL benchmarks and discovered that GA can compete with popular RL algorithms such as A3C \cite{MnihBMGLHSK16}, DQN, and the OpenAI-ES. GAs also possess the same scalability advantage as ES, that can drastically speedup run time if distributed computing resources are available. Gangwani and Peng \cite{GangwaniP18} introduced a new GA with imitation learning for policy crossovers in state space, producing offspring policies that effectively mimic their best parent in generating similar state visitation distributions. This idea was shown to lessen catastrophic performance drops stemming from naive GA crossovers in parameter space---which tend to destroy the hierarchical relationship of the networks.

In this paper, we propose an alternative approach to data-efficient evolutionary RL that reduces  expensive  agent-environment interaction data requirements. Our method exploits the temporal structure of RL with auxiliary tasks of progressive episode lengths. Different from \cite{FuksAHL19}, our algorithm is the first to augment ES with skills transfer by means of evolutionary multitasking. Preliminaries of these algorithmic components are discussed next.

\section{Preliminaries}
In this section, we first present the general problem statement for direct policy search in RL. The OpenAI-ES \cite{SalimansHCS17}, which forms the base algorithm for NuEMT, is described. The basics of EMT and its probabilistic formulation are presented next. The section ends with a short overview of importance sampling for probabilistic inference.

\subsection{Direct Policy Search}
In RL, the goal is to find a policy, i.e., a state-action mapping function, that maximizes cumulative rewards over time of an agent operating in a dynamic and uncertain environment. The policy determines how the agent interacts with the environment. The reward maximization problem can be cast as one of policy parameter optimization as:
\begin{equation}\label{eq:objective_function}
    \max_{\theta \in \mathbb{R}^{n}} F(\pi_{\theta}),
\end{equation}
\noindent where the objective (aka fitness) function $F$ is the total returns achieved under policy $\pi_{\theta}$, parameterized by $\theta$.

There are several methods for solving the maximization problem in Eq. (\ref{eq:objective_function}), such as policy iteration \cite{Bertsekas05}, policy gradients \cite{silver2014deterministic}, or derivative-free optimization \cite{ManiaGR18}. In this work, we focus on derivative-free ES for training neural network policies (where $\theta$ represents the weights of the network), thus allowing $F$ to be treated as a black-box without restriction on the distribution of rewards (sparse or dense), etc.

\subsection{ES for Direct Policy Search}\label{section3.b}
Consider the OpenAI-ES, an ES variant that belongs to the class of \emph{natural evolution strategies} \cite{SalimansHCS17, WierstraSGSPS14}. The algorithm is based on adding isotropic Gaussian noise of covariance $\sigma^2 I$ to a mean vector $\tilde{\theta}$, transforming Eq. (\ref{eq:objective_function}) to the following maximization of the expected reward (averaged over the induced probability distribution in parameter space):
\begin{equation}\label{eq:gaussian_smoothed_objective}
     \max_{\tilde{\theta} \in \mathbb{R}^{n}} \mathbb{E}_{\theta \sim \mathcal{N}(\tilde{\theta}, \sigma^2 I)}[F(\pi_{\theta})].
\end{equation}
\noindent This Gaussian-blurred version of the objective function helps to remove non-smoothness introduced by the environment, hence enabling $\tilde{\theta}$ to be effectively updated by the following expected reward gradient (derived by the `log-likelihood trick'):
\begin{equation}\label{eq:gradient_estimate}
    \nabla_{\tilde{\theta}} \mathbb{E}_{\theta \sim \mathcal{N}(\tilde{\theta}, \sigma^2 I)} [F(\pi_{\theta})] = \frac{1}{\sigma^2} \mathbb{E}_{\theta}[F(\pi_{\theta}) (\theta-\tilde{\theta})].
\end{equation}
\noindent In practice, the gradient is approximated via the Monte Carlo method. A fixed number of samples---equivalent to the population size $N$ of the ES algorithm---are drawn from $\mathcal{N}(\tilde{\theta},\sigma^2I)$ to compute stochastic gradient estimates of the policy update in every iteration. A pseudocode of the OpenAI-ES is shown in Algorithm \ref{algo:OpenAI-ES}. The algorithm first perturbs parameter vector $\tilde{\theta}$ by sampling $\epsilon_i$'s from a multivariate normal distribution $\mathcal{N}(0, I)$; see steps 3 and 4. The perturbed parameter values $\theta_i = \tilde{\theta} + \sigma\epsilon_i$ are then evaluated by running an episode in the environment with the corresponding policy. The results ($F_i$'s) obtained from these episodes approximate the gradient in Eq. (\ref{eq:gradient_estimate}) as $\frac{1}{N\sigma^2} \sum_{i=1}^{N} F_i * (\theta_i - \tilde{\theta}_t)$, which is then used to update the mean $\tilde{\theta}$. The above repeats until a terminal condition (e.g., function evaluation budget) is met.

\begin{algorithm}[!htb]
\caption{Pseudocode of the OpenAI-ES}\label{algo:OpenAI-ES}
\begin{algorithmic}[1]
    \INPUT $\alpha$: step size, $\sigma$: noise standard deviation, $\tilde{\theta}_0$: initial policy parameters, $N$: population size
    \State Set $t=0$
    \Repeat
        \State Sample $\epsilon_1, \epsilon_2, \ldots, \epsilon_{N} \sim \mathcal{N}(0, I)$
        \State Let $\theta_i = \tilde{\theta}_t + \sigma\epsilon_i$
        \State Collect $N$ returns, $F_i = F(\pi_{\theta_i}), \forall i\leq N$
        \State Update $\tilde{\theta}_{t+1} \leftarrow \tilde{\theta}_{t} + \frac{\alpha}{N\sigma^2} \sum_{i=1}^{N} F_i * (\theta_i - \tilde{\theta}_t)$ 
        \State Set $t = t+1$
    \Until{\textit{termination condition is met}}
    
\end{algorithmic}
\end{algorithm}

A parallel implementation of the OpenAI-ES was discussed in \cite{SalimansHCS17} where evaluations were handled independently among distributed workers. The main novelty there was that the algorithm made use of shared random seeds, reducing the bandwidth required for
communication between the workers. Such implementations can greatly reduce run time, making ES comparable with other hardware-accelerated approaches such as the training of deep RL agents with GPUs.

\subsection{Basics of Evolutionary Multitasking}

The motivation behind EMT is to enhance evolutionary search by the exchange and reuse of evolved skills between jointly optimized tasks. This idea can be applied to boost convergence rates in a difficult target task by solving it in tandem with a group of related \emph{auxiliary tasks} that are simpler and/or of lower computational cost. Successful applications of this type with EMT have been reported in the literature \cite{Nick2021, ma2021enhanced, ding2017generalized}. Specifically, the auxiliary tasks serve as informative proxies that quickly guide the target optimization process towards promising regions of the search/parameter space, by the adaptive transfer of discovered solutions.

Let us consider a scenario with $K$ optimization tasks solved simultaneously. Suppose, without loss of generality, each task $T_i$, $\forall i \in \{1, 2, \dots, K\}$, to be a maximization problem instance with search space $\Theta_i$ and objective function $F_i : \Theta_i \rightarrow \mathbb{R}$. Each task may be subject to additional equality and/or inequality constraints. In this setting, the goal of a multitask algorithm is to find in a single run a \emph{set} of optimal solutions $\{\theta^*_1, \theta^*_2, \ldots, \theta^*_K\} = \argmax \{F_1(\theta), F_2(\theta), \ldots, F_K(\theta)\}$, such that $\theta^*_i \in \Theta_i$ and satisfies all constraints of $T_i$, $\forall i$. 

In many examples of EMT, the tasks are defined in the same search space \cite{OngG16}, i.e., $\Theta_1 = \Theta_2 = \ldots = \Theta_K$, while their objective functions $F_1, F_2, \ldots, F_K$ may differ. (This is also true in the present paper, since all tasks are defined in a common space of neural network policy parameters.) In such cases, we symbolize the single \emph{unified space}, encompassing all task-specific search spaces, simply as $\Theta$. The unified space provides a shared pathway for distinct but possibly related tasks to exchange mutually beneficial information. For instance, the direct transfer of elite solutions between tasks with correlated objective functions could lead to the rapid discovery of performant solutions. Substantial speedups can thus be achieved in comparison to conventional methods that re-explore search spaces from scratch. 

In this paper, we consider a special case of EMT where we are primarily interested in solving a target task denoted hereinafter as $T_K$. The remaining $T_1, T_2, \ldots, T_{K-1}$ act as auxiliary tasks catalysing the evolutionary search. 

\subsection{A Probabilistic View of Evolutionary Multitasking}\label{section:3.d}

Let the target optimization task $T_K$ be:
\begin{equation}\label{eq:generic_max_eqn}
    \max_{\theta \in \Theta_K} F_K(\theta).
\end{equation}

\noindent Through the lens of probabilistic model-based evolutionary search, we may reformulate Eq. (\ref{eq:generic_max_eqn}) as \cite{WierstraSGSPS14},

\begin{equation}\label{eq:probabilistic_max_eqn}
    T_K : \max_{p_K(\theta)} \int_{\Theta_K} F_K(\theta) \; p_K(\theta) \; d\theta,
\end{equation}

\noindent where $p_K(\theta)$ represents the probability distribution of an evolving population of candidate solutions. Notice that if $\theta_K^*$ maximizes $F_K(\theta)$, i.e., $F_K(\theta_K^*) = \max_{\theta \in \Theta_K} F_K(\theta)$, then the optimal probability distribution model $p^*_K(\theta)$ that solves Eq. (\ref{eq:probabilistic_max_eqn}) is given by $\delta(\theta - \theta_K^*)$, where $\delta(\theta - \theta_K^*)$ is a Dirac delta function centred at $\theta_K^*$. This result follows from the identity: $\int_{\Theta_K} F_K(\theta) \; \delta(\theta - \theta_K^*) \; d\theta = F_K(\theta_K^*)$. As such, it is observed that probabilistic reformulation does not change the eventual outcome of the optimization problem.

In the EMT setting, the individual probabilistic models $p_1(\theta)$, $p_2(\theta)$, $\dots$, $p_{K-1}(\theta)$ pertaining to $T_1$, $T_2$, $\dots$, $T_{K-1}$, respectively, are accessible to $T_K$ whilst being jointly optimized in unified space $\Theta$. The individual models encode the skills evolved for the different tasks. Hence, we seek to activate these additional building-blocks of knowledge to accelerate the target search. To that end, we further generalize the probabilistic reformulation of Eq. (\ref{eq:probabilistic_max_eqn}) by defining a mixture model (in unified space $\Theta$) as,
\begin{equation}\label{eq:mixture_model_max_eqn}
    T_K : \max_{w_{K,1}, \ldots, w_{K,K}, p_K(\theta)} \int_{\Theta} F_K(\theta) \sum_{i=1}^{K} w_{K,i} \cdot p_i(\theta) \; d\theta,
\end{equation}
\noindent where $w_{K,i \neq K} \geq 0$, $w_{K,K} > 0$ and $\sum_{i=1}^{K} w_{K,i} = 1$. Eq. (\ref{eq:mixture_model_max_eqn}) is the fundamental equation underpinning our proposed NuEMT algorithm (to be fully developed in Section IV). Note that the generalized formulation is optimized if $p^*_K(\theta) = \delta(\theta - \theta_K^*)$ and $w_{K,K} = 1$, implying that the outcome of optimization still remains unchanged under Eq. (\ref{eq:mixture_model_max_eqn}). By setting $w_{K,K} = 1$ and $w_{K,i \neq K} = 0$, indicating zero inter-task transfers, the mixture model simply collapses to $p_K(\theta)$, and we fall back to the well known form of Eq. (\ref{eq:probabilistic_max_eqn}). 

However, leveraging the mixture model opens new pathways to actualize skills transfer in EMT, through solution cross-sampling. The mixture coefficient $w_{K, i} \; \forall i$ essentially reflects the transferrability of solutions from a \emph{source} task $T_i$ to the target $T_K$. Precisely, if candidate solutions evolved for the \textit{i}th task---i.e., drawn from the probabilistic model $p_i(\theta)$---are found to be performant (i.e., return high rewards) for $T_K$ as well, then the value of $w_{K,i}$ can be increased to intensify the cross-sampling of solution prototypes. In contrast, if solutions transferred from a given source do not excel at $T_K$, then the corresponding mixture coefficient can be gradually neutralized. A detailed discussion on this topic can be found in a recent survey \cite{gupta2022half}.

\subsection{Probabilistic Inference by Importance Sampling}\label{overviewIS}

Importance sampling is a general statistical technique for inferring properties of a nominal probability distribution $p(\theta)$, given samples drawn from a different distribution $q(\theta)$. The technique is widely used for variance reduction in Monte Carlo methods where $q(\theta)$ takes the form of a biasing distribution from which samples ($\theta's$) are drawn; an instructive review on the subject can be found in \cite{tokdar2010importance}, and applications in RL can be referred in \cite{hanna2021importance}. In this paper, we shall utilize importance sampling in a unique manner to induce inter-task skills transfer in the parameter space of EMT. 

Let $\mathbb{E}_{p_i(\theta)}[F_i(\theta)]$ be the expectation of $F_i(\theta)$ under the nominal distribution $p_i(\theta)$ in the parameter space $\Theta_i \subset \mathbb{R}^n$. If the biasing distribution $q_i(\theta)$ is also defined in $\mathbb{R}^n$ such that $supp(p_i) \subseteq supp(q_i)$, where $supp(p_i) = \{\theta : p_i(\theta) > 0\}$, then the expectation $\mathbb{E}_{p_i(\theta)}[F_i(\theta)]$ can be reformulated as:
\begin{equation}\label{impsamp}
\begin{split}
    \mathbb{E}_{p_i(\theta)}[F_i(\theta)]
    & = \int_{\Theta_i} F_i(\theta) p_i(\theta) \; d\theta \\
    & = \int_{\Theta_i} \frac{F_i(\theta)p_i(\theta)}{q_i(\theta)} q_i(\theta) \; d\theta \\
    & = \mathbb{E}_{q_i(\theta)}\left[\frac{F_i(\theta)p_i(\theta)}{q_i(\theta)}\right].
\end{split}
\end{equation}
Here, the multiplicative adjustment to $F_i(\theta)$ given by importance weights $p_i(\theta)/q_i(\theta)$ compensates for sampling from a different distribution $q_i(\theta)$ whilst inferring properties of the nominal distribution $p_i(\theta)$. In the context of EMT (as presented in Section \ref{section:3.d}), Eq. (\ref{impsamp}) suggests a technique for updating the probabilistic model $p_K(\theta)$ of target task $T_K$ using solution samples from a different probability distribution (namely, the mixture $\sum_{i=1}^{K} w_{K,i} \cdot p_i(\theta)$ in Eq. (\ref{eq:mixture_model_max_eqn})); hence, leading to the transfer of information through cross-sampling. This new insight lies at the core of the proposed NuEMT algorithm developed in the next section. 

\section{NuEMT with Importance Sampling}
\begin{figure}[!htb] 
\centering
\includegraphics[width=0.49\textwidth]{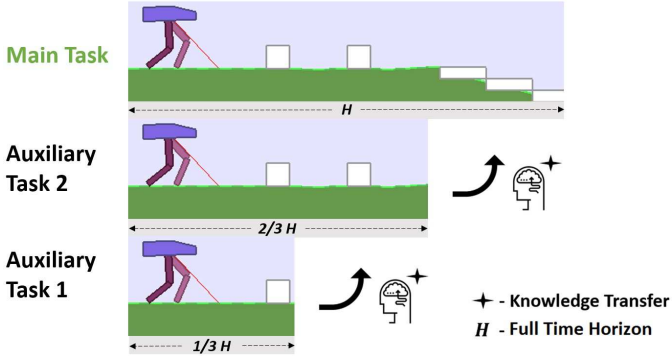}
\caption{Illustrating two auxiliary tasks of shorter episode length ($H/3$ and $2H/3$, respectively) extracted from the main target task (of episode length $H$). Useful skills that are quickly evolved in shorter episodes are progressively transferred to longer and harder tasks (e.g., those with more obstacles).}
\label{fig:nuemt_diagram}
\end{figure}

In this section, we first describe the creation of the auxiliary tasks that catalyse policy search in the target task at hand. Next, we describe the NuEMT algorithm with importance sampling for skills transfer. An adaptive resource allocation strategy is proposed to dynamically adjust computational resources assigned to each constituent task in NuEMT. 

\subsection{Construction of Auxiliary Tasks}\label{section:4.a}

Skills can be seen as the culmination of a continuous learning process, through accumulating and building on experiences gained from every long or short training session. Bringing this perspective to RL, it is believed that the transfer of skills from simpler tasks (of shorter episode length) could help solve longer and harder tasks more effectively. If tasks with shorter episodes are extracted from the longer and harder task, underlying similarities are expected to exist between them.

In this paper, we realize the aforementioned idea for the first time in a multitask setting. To guide the policy search of the target task $T_K$---of maximum episode length `$H$'---we construct a set of auxiliary tasks, $T_1$, $T_2$, $\ldots$, $T_{K-1}$, with shorter episodes of progressive lengths to be solved jointly with $T_K$. \emph{The design of the NuEMT algorithm is such that solutions encoding skills evolved in tasks with shorter episodes are progressively transferred to those that are longer and harder}. A visualization of the idea is provided in Figure \ref{fig:nuemt_diagram}, where there is a total of $K = 3$ tasks, i.e., two auxiliary tasks and one main task. We standardize the episode length of the auxiliary tasks according to their indices, i.e., the $i$th auxiliary task has an episode length of $\frac{i}{K} \cdot H$. 

Given this setup, the probabilistic formulation for $T_i$ can be written in terms of a mixture model (similarly to Eq. (\ref{eq:mixture_model_max_eqn})) as:
\begin{equation}\label{eq:mixture_model_objective_function}
    \max_{p_i({\theta}), w_{i, 1}, \ldots, w_{i,i}} \mathbb{E}_{q_i({\theta})} [F_i(\pi_\theta)] = \int_{\Theta} F_i(\pi_{\theta}) \; q_i(\theta) \; d\theta,
\end{equation}

\noindent where $p_i(\theta)$ is the $i$th task-specific probabilistic model, and $q_i(\theta) = \sum_{j \leq i} w_{i,j} \cdot p_j(\theta)$ represents the assimilated mixture model of $T_i$. Note that when $i=1$, the mixture reduces as $q_1(\theta) = p_1(\theta)$. Hence, no knowledge transfer occurs to the task $T_1$ with shortest episode length.

\subsection{Naive Stochastic Gradient Estimates}\label{section:4.b}
From Section \ref{section3.b}, recall the use of an isotropic multivariate Gaussian, parameterized by its mean $\tilde{\theta}$, as the search distribution in the OpenAI-ES. Accordingly, for task $T_i$ we define $p_i(\theta) = \mathcal{N}(\tilde{\theta_i}, \sigma^2 I)$. The gradient $\nabla_{\tilde{\theta}_{i}}$ of the objective function in Eq. (\ref{eq:mixture_model_objective_function}) is:

\begin{equation}\label{eq:theta_update_without_is}
\begin{split}
    \nabla_{\tilde{\theta}_{i}} \mathbb{E}[F_i(\pi_{\theta})] & = \nabla_{\tilde{\theta}_{i}} \int_{\Theta} F_i(\pi_{\theta}) \; q_i(\theta) \; d\theta \\
    & = w_{i,i} \int_{\Theta} F_i(\pi_{\theta})\; p_i(\theta) [\nabla_{\tilde{\theta}_{i}} \log p_i(\theta)] \; d\theta,
\end{split}
\end{equation}
where the second statement comes from the log-likelihood trick. Eq. (\ref{eq:theta_update_without_is}) yields the Monte Carlo approximation:
\begin{equation}\label{eq:theta_update_without_is_2}
\begin{split}
    \nabla_{\tilde{\theta}_{i}} \mathbb{E}[F_i(\pi_{\theta})]
    & \approx \frac{1}{N_i} \frac{w_{i,i}}{\sigma^2} \sum_{k=1}^{N_i} F_i(\pi_{\theta_k}) (\theta_k - \tilde{\theta}_i).
\end{split}
\end{equation}

\noindent Here, $\theta_k \sim p_i(\theta)$ and $N_i$ is the total samples (i.e., population size) assigned to task $T_i$. As per Eq. (\ref{eq:theta_update_without_is_2}), it is clear that solutions sampled from probabilistic models of tasks $T_{j \neq i}$ would not directly exert any influence on the stochastic gradient updates of $p_i(\theta)$. In other words, skills transfer between tasks is non-existent in this naive approach. Therefore, a modification to the gradient estimation is needed for inter-task transfers to be established. This can be achieved by means of importance sampling, as disclosed next.

\subsection{Importance Sampling for Skills Transfer}\label{section:4.c}

Note that the mixture model $q_i(\theta)$ in Eq. (\ref{eq:mixture_model_objective_function}) includes the task-specific model $p_i(\theta)$ as one of its components. With $w_{i,i} > 0$, the condition $supp(p_i) \subseteq supp(q_i)$ for importance sampling (see Section \ref{overviewIS}) is satisfied. Thus, we can estimate expectations under $p_i(\theta)$---as in Eq. (\ref{eq:theta_update_without_is})---by using samples from the mixture model $q_i(\theta)$ instead. Accordingly, rewriting $p_i(\theta)$ as $\frac{p_i(\theta)}{q_i(\theta)} q_i(\theta)$ and plugging this into Eq. (\ref{eq:theta_update_without_is}), we get:

\begin{equation}\label{eq:theta_update}
\begin{split}
    \nabla_{\tilde{\theta}_{i}} \mathbb{E}[F_i(\pi_{\theta})] & = w_{i,i} \int_{\Theta} F_i(\pi_{\theta}) \; \frac{p_i(\theta)}{q_i(\theta)} \; [\nabla_{\tilde{\theta}_{i}} \log p_i(\theta)] \;q_i(\theta)\; d\theta \\
    & \approx \frac{w_{i,i}}{N_i\sigma^2} \sum_{k=1}^{N_i} F_i(\pi_{\theta_k}) \; \frac{p_i(\theta_k)}{\sum\limits_{l=1}^i w_{i,l} \cdot p_l(\theta_k)} (\theta_k - \tilde{\theta}_i),
\end{split}
\end{equation}

\noindent where $\theta_k \sim q_i(\theta)$. As a result of the reformulation, solutions sampled from all components of the mixture model $q_i(\theta)$ shall directly influence stochastic gradient updates of the search distribution parameter $\tilde{\theta}_i$ of $T_i$. Hence, through importance sampling, the transfer of solution prototypes encoding evolved skills is activated from tasks of shorter episode lengths to those that are longer and harder.

In addition to $\tilde{\theta}_i$, the mixture coefficients of $q_i(\theta)$ are also updated during the search process. Applying importance sampling again, the gradient estimate with respect to $w_{i,j}$ ($j\leq i$) can be obtained as:
\begin{equation}\label{eq:mixture_weight_update}
\begin{split}
    \frac{\partial \mathbb{E}[F_i(\pi_{\theta})]}{\partial w_{i,j}}
    & = \int_{\Theta} F_i(\pi_{\theta}) \; p_j(\theta) \; d\theta \\ & = \int_{\Theta} F_i(\pi_{\theta}) \; \frac{p_j(\theta)}{q_i(\theta)} q_i(\theta) \; d\theta \\
    & \approx \frac{1}{N_i} \sum_{k=1}^{N_i} F_i(\pi_{\theta_k}) \; \frac{p_j(\theta_k)}{\sum\limits_{l=1}^i w_{i,l} \cdot p_l(\theta_k)}.
\end{split}
\end{equation}

\subsection{Derived NuEMT Update Equations}
In order to make the methodology robust and invariant to outliers as well as arbitrary yet order-preserving fitness transformations, a standard rank-based fitness shaping function defined by utility values $u_i$ as substitute to the actual returns $F_i$ is considered \cite{WierstraSGSPS14};
\begin{equation}
    u_{i,k} = \frac{max(0, \log (\frac{N_i}{2}+1) - \log k)}{\sum_{n=1}^{N_i} max(0, \log (\frac{N_i}{2}+1) - \log n)},
\end{equation}

\noindent where $u_{i,k}$ is the utility of the $k$th sample in a \emph{sorted} population list, i.e., $F_i(\pi_{\theta_1}) \geq F_i(\pi_{\theta_2}) \geq \ldots \geq F_i(\pi_{\theta_k}) \geq \ldots \geq F_i(\pi_{\theta_{N_{i}}})$, in the population of $T_i$.

In practice, if the gap between $\tilde{\theta_j}$ and $\tilde{\theta_i}$ becomes large, the importance weight $\frac{p_i(\theta_k)}{q_i(\theta_k)}$ in Eq. (\ref{eq:theta_update}) rapidly approaches zero for $\theta_k \sim p_j(\theta)$ due to distribution sparsity in even moderately high dimensional parameter spaces. The importance weights thus suppress the influence of cross-sampled solutions on the update of $\tilde{\theta_i}$. To resolve this issue, we adopt a projection technique introduced in \cite{WongGO21}; accordingly, solutions $\theta_k \sim p_{j}(\theta)$ that lie outside Mahalanobis distance $r$ from $p_i(\theta)$ are mapped back to a distance $r$ (set to 1 in all experiments) while maintaining the same directional bias. A mapped solution, $\theta^{'}_k$, is defined as:
\begin{equation}
{\footnotesize \theta^{'}_k =} {\footnotesize\begin{cases} \tilde{\theta_i} + (\theta_k - \tilde{\theta_i}) \cdot \min \left(1, \frac{r}{\norm{\frac{\theta_k - \tilde{\theta_i}}{\sigma}}}\right)  &\text{if $\theta_k \not\sim p_{i}(\theta)$}\\
\theta_k &\text{otherwise},
    \end{cases}}
\end{equation}
with its utility value simply retained as $u_{i,k}(\pi_{\theta^{'}_k}) \leftarrow u_{i,k}(\pi_{\theta_k})$.

Based on the above, the final update equation for parameter $\tilde{\theta_i}$ is given by:
\begin{equation}\label{theta_update}
    \tilde{\theta}_i \leftarrow \tilde{\theta}_i + \frac{\alpha}{N_i} \frac{w_{i,i}}{\sigma^2} \sum_{k=1}^{N_i} u_{i, k} \; \frac{p_i(\theta^{'}_k)}{q_i(\theta^{'}_k)} (\theta^{'}_k - \tilde{\theta}_i),
\end{equation}
\noindent where $\alpha$ is the learning stepsize for $\tilde{\theta_i}$. 

For the update of the mixture coefficients, the constraints $\sum_{j \leq i} w_{i,j} = 1$ and $w_{i,j} \geq 0$ must be satisfied. Thus, an additional step projecting the gradient approximation of Eq. (\ref{eq:mixture_weight_update}) to the constraint plane $C$ is needed. Denoting the normal vector to the plane as $\vec{c} = [1, \ldots, 1]$, and the gradient estimate as $\vec{b} =  [\frac{\partial \mathbb{E}[F_i(\pi_{\theta})]}{\partial w_{i,1}}, \frac{\partial \mathbb{E}[F_i(\pi_{\theta})]}{\partial w_{i,2}}, \ldots, \frac{\partial \mathbb{E}[F_i(\pi_{\theta})]}{\partial w_{i,i}}]$, the scaled projection of the gradient on the plane takes the form:
\begin{equation}\label{eq:projection}
    proj_{C}(\vec{b}) = \beta \cdot (\vec{b} - \frac{\vec{b} \cdot \vec{c}}{i}\vec{c}),   
\end{equation}
\noindent where $\beta$ is the learning stepsize for the mixture coefficients. Then, the final update equation for $\mathbf{w}_i = [w_{i,1}, w_{i,2}, \ldots, w_{i,i}]$ takes the following form:
\begin{equation}\label{eq:weights_update}
    \mathbf{w}_i \leftarrow \mathbf{w}_i + \lambda \cdot proj_{C}(\vec{b}),
\end{equation}
\noindent where $\lambda > 0$ is a scaling factor to ensure nonnegativity. 
It can be shown that the coefficients converge to steady values under Eq. (\ref{eq:weights_update}). Assume (for simplicity) the individual probabilistic models $p_1(\theta)$, $p_2(\theta)$, $\dots$, $p_{i}(\theta)$ pertaining to $T_1$, $T_2$, $\dots$, $T_{i}$ have converged to the optimal Dirac delta functions $\delta(\theta - \theta_1^*)$, $\delta(\theta - \theta_2^*)$, $\dots$, $\delta(\theta - \theta_i^*)$, respectively, with $\theta_i^* \neq \theta_j^*$, $\forall j \neq i$. From the first line of Eq. (\ref{eq:mixture_weight_update}), it follows that $\frac{\partial \mathbb{E}[F_i(\pi_{\theta})]}{\partial w_{i,i}} > \frac{\partial \mathbb{E}[F_i(\pi_{\theta})]}{\partial w_{i,j}}$, since $\int_{\Theta} F_i(\theta) \; \delta(\theta - \theta_i^*) \; d\theta > \int_{\Theta} F_i(\theta) \; \delta(\theta - \theta_j^*) \; d\theta$. Plugging the inequality into Eq. (\ref{eq:projection}), the $i$th component of $proj_{C}(\vec{b})$ is:

\begin{equation}
\begin{split}
     \beta \cdot \left(\frac{\partial \mathbb{E}[F_i(\pi_{\theta})]}{\partial w_{i,i}} - \frac{1}{i}\sum_{j=1}^i \frac{\partial \mathbb{E}[F_i(\pi_{\theta})]}{\partial w_{i,j}}\right) > 0.
\end{split}
\end{equation}

\noindent This shows that the mixture coefficient $w_{i, i}$ tends to increase under the update. What's more, the scaling factor $\lambda$ in Eq. (\ref{eq:weights_update}) serves to ensure $w_{i, i} \leq 1$. Hence, $w_{i, i}$, $\forall i$, gradually converges to a steady value of 1, while all other $w_{i, j}$'s go to 0. We demonstrate this behavior through an experimental result in Section V-D of the paper.

\subsection{Adaptive Resource Allocation Strategy}\label{section:4.d}

Learning a high mixture coefficient $w_{K,j}$ suggests strong transferrability between $T_K$ and $T_j$. The auxiliary task $T_j$ could then serve as a cheap proxy for the main task $T_K$ for quickly progressing the target search. Greater computational resources can be allocated to $T_j$, while facilitating the transfer of evolved skills to $T_K$ at a fraction of the cost. Such a resource allocation strategy has already been demonstrated to work well in the context of EMT for evolutionary machine learning \cite{Nick2021}. With that in mind, we incorporate an adaptive resource allocation strategy into NuEMT, assigning population sizes for the evolution of the auxiliary tasks as:

\begin{equation}\label{resources}
    N_j = N_{total} \cdot w_{K,j},
\end{equation}
\noindent where $N_{total}$ is the total population size of the NuEMT, and $N_j$ is the size allocated to $T_j$. Eq. (\ref{resources}) reduces resource wastage, since an auxiliary task uncorrelated with the target will get a small or zero population size. 

\begin{algorithm}[!htb]
\caption{Pseudocode of the NuEMT algorithm}\label{algo:NuEMT}
\begin{algorithmic}[1]
\INPUT $N_{total}$: population size, $K$: number of tasks, $\sigma$: noise standard deviation, $\alpha$: stepsize, $\beta$: mixture stepsize, $H$: full time horizon, $\mathcal{T} = \{T_1, T_2, \ldots, T_K\}$: set of tasks

\State Set mixture coefficient $w_{i,j}$ = $1/i$, $\forall j \leq i$ for $T_i, \; \forall i$
\State Set population size $N_{i}$ = $w_{K, i} \cdot N_{total}$ for $T_i, \; \forall i$
\State Set $H_i = \frac{i}{K} \cdot H, \forall i \leq K$
\State Set $\tilde{\theta}_{i} = \textbf{0} \in \mathbb{R}^{n}$ for $T_i, \; \forall i$
\Repeat
    \ForEach {$T_i \in \mathcal T$}
        \State\multiline{Set $q_i(\theta) = \sum_{j \leq i} w_{i,j} \cdot p_{j}(\theta)$, where $p_{j}(\theta)= \mathcal{N}(\tilde{\theta}_j, \sigma^2 I)$}
        \State Sample $\theta_1, \theta_2, \ldots, \theta_{N_{i}} \sim q_i(\theta) $
        \State Collect $N_i$ based on episode length $H_i$:
        \State 
            \begin{equation*}
                F_{i,k} = F_i(\pi_{\theta_k}), \forall k \leq N_i 
            \end{equation*}
        \State\multiline{Sort $F_{i,k}$ in descending order of fitness, and apply rank-based fitness shaping:}
        \State
            \begin{equation*}
                u_{i,1} \geq \ldots \geq u_{i,N_i} \Leftarrow F_{i,1} \geq \ldots \geq F_{i, N_i}
            \end{equation*}
        \State\multiline{Project solutions not sampled from $p_i(\theta)$:}
        {\footnotesize\State\multiline{\[\theta^{'}_k = \begin{cases} \tilde{\theta_i} + (\theta_k - \tilde{\theta_i}) \cdot \min \left(1, \frac{r}{\norm{\frac{\theta_k - \tilde{\theta_i}}{\sigma}}}\right) &\text{if } \theta_k \not\sim p_{i}(\theta)\\
        \theta_k &\text{otherwise} \end{cases} \]}}
        \State\multiline{Calculate gradient estimates for $\theta_{i}$ and $w_{i,j}$:}
        \State
            \begin{equation*}
            \begin{split}
                &\nabla_{\theta_i} = \frac{1}{{N_i}}\frac{w_{i,i}}{\sigma^2} \sum_{k=1}^{N_i} u_{i,k} \cdot \frac{p_{i}(\theta^{'}_k)}{q_i(\theta^{'}_k)} (\theta^{'}_k - \tilde{\theta}_i)
                \\
               &\nabla_{w_{i,j}} = \frac{1}{N_i} \sum_{k=1}^{N_i} u_{i,k} \cdot \frac{p_{j}(\theta_k)}{q_i(\theta_k)}, \forall j \leq i
            \end{split}
            \end{equation*}
        \State Perform update step:
        \State 
            \begin{equation*}
                \begin{split}
                    &\tilde{\theta}_i \leftarrow \tilde{\theta}_i + \alpha \cdot \nabla_{\theta_i}
                    \hspace{-15mm}
                    \\
                    &\mathbf{w}_{i} \leftarrow \mathbf{w}_{i} + \lambda \cdot proj_{C}([\nabla_{w_{i,1}}, \nabla_{w_{i,2}}, \ldots, \nabla_{w_{i,i}}]) 
                    \\
                \end{split}
            \end{equation*}
    \EndFor
    \State Update $N_{i} \leftarrow w_{K, i} \cdot N_{total}, \forall i \leq K$
\Until{\textit{termination condition is met}}
\end{algorithmic}
\end{algorithm}

\subsection{Summarizing NuEMT for RL}
Tying together the derived update equations, we herein summarize the NuEMT algorithm for RL. Our methodology incorporates mixture modelling as a means of inter-task relationship capture, to control the \emph{extent} of skills transfer between tasks. Auxiliary task $T_i$, for $i = 1,2,\ldots,K-1$, is extracted from the same environment as the main task $T_K$, but has a shorter episode length, i.e., $\frac{i}{K} \cdot H$. A pseudocode of the overall procedure is given in Algorithm \ref{algo:NuEMT}.  

At initialization, all components of the mixture models are uniformly weighted by setting $w_{i,j} = 1/i$ for $j = 1, 2, \ldots, i$. An equal population size of $N_i = \frac{N_{total}}{K}$ is allocated to each task. In each iteration, every task samples solutions, $\theta_1, \theta_2, \ldots, \theta_{N_i}$ from its mixture model $q_i(\theta)$. Each $\theta_k$ represents a policy $\pi_{\theta_k}$, and $F_i(\pi_{\theta_k})$ represents the total reward received from an episode run with length $H_i = \frac{i}{K} \cdot H$. The $N_i$ reward values received are then used for parameter updates as formulated in Eq. (\ref{theta_update}) and Eq. (\ref{eq:weights_update}). Subsequently, the coefficients $w_{K,\cdot}$ from the main task's mixture model $q_K(\theta)$ are used to determine the population size to be allocated to each of the auxiliary tasks in the next iteration. This process continues until a terminal condition is met.

In our implementation, we find it useful to perform state normalization \cite{ManiaGR18} as it enables different state components to have a fair share of influence during training. A similar normalization approach known as \emph{virtual batch normalization} is also used by OpenAI-ES \cite{SalimansHCS17}. In addition, weight decay is added as a form of regularization to prevent parameters of the policy network from exploding. Lastly, we adopt \emph{mirror sampling} \cite{BrockhoffAHAH10} as a variance reduction technique.

It is worth noting that when we compare NuEMT with a conventional neuroevolutionary algorithm, the computational cost of NuEMT (per iteration) will be lower given the same $N_{total}$. This is because each of the $K$ tasks is assigned $\frac{N_{total}}{K}$ solutions to start with, and, assuming the computational cost of evaluating a solution for $T_i$ to be $C_i$, the total cost will be $\sum_{i} C_i \cdot \frac{N_{total}}{K} < N_{total} \cdot C_K$ since $C_1 < C_2 < \ldots < C_K$. This is especially crucial for RL problems that may be dealing with extremely long episodes in large-scale simulations.

\section{Experimental Studies}
\begin{figure}[!htb]
\minipage[t]{0.315\linewidth}
    \includegraphics[height=2cm, width=1\textwidth]{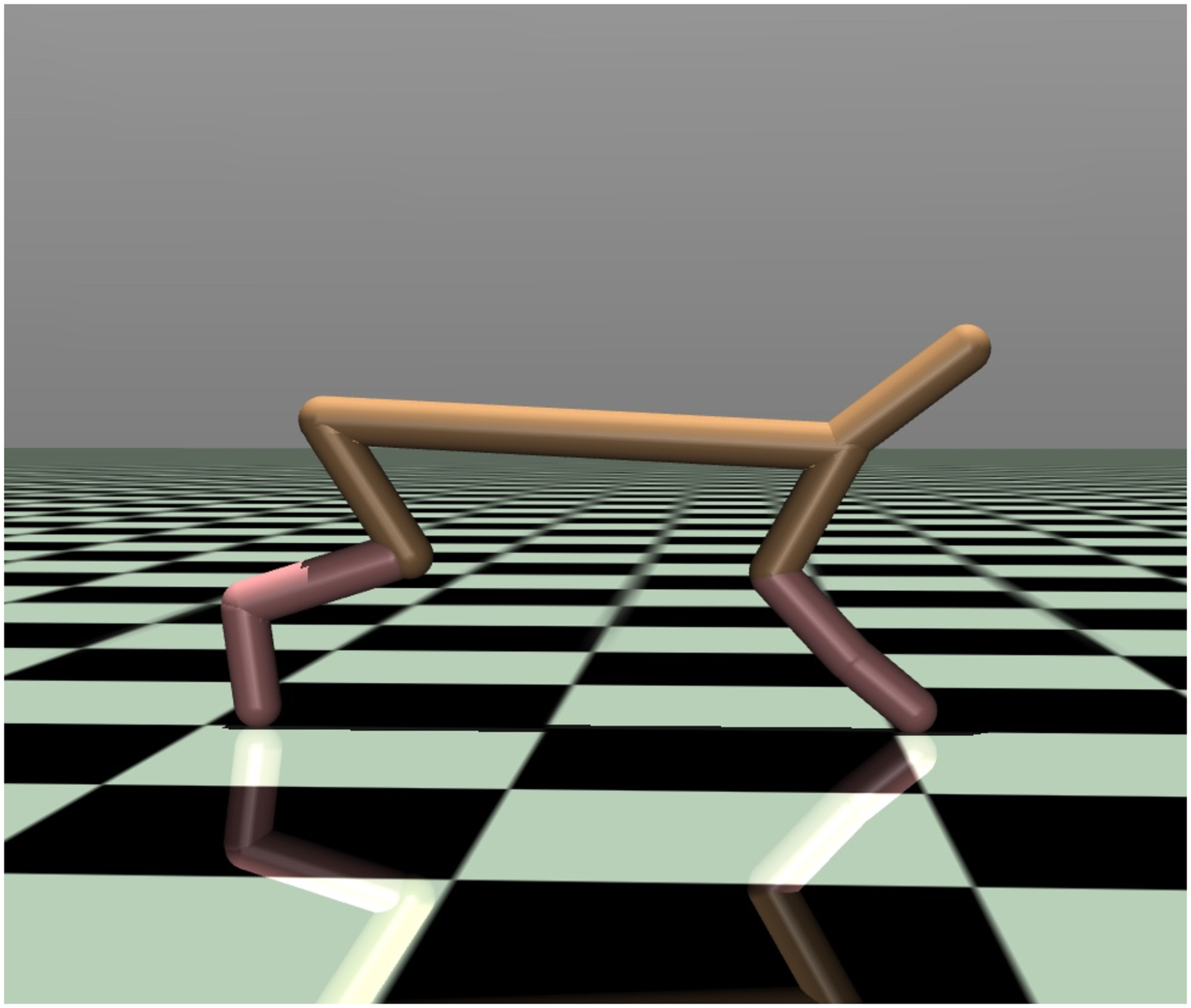}
    \centering
    \subcaption{HalfCheetah-v2}
    \label{fig:HalfCheetah-v2_image}
\endminipage\hfill
\centering
\minipage[t]{0.315\linewidth}
    \includegraphics[height=2cm, width=1\linewidth]{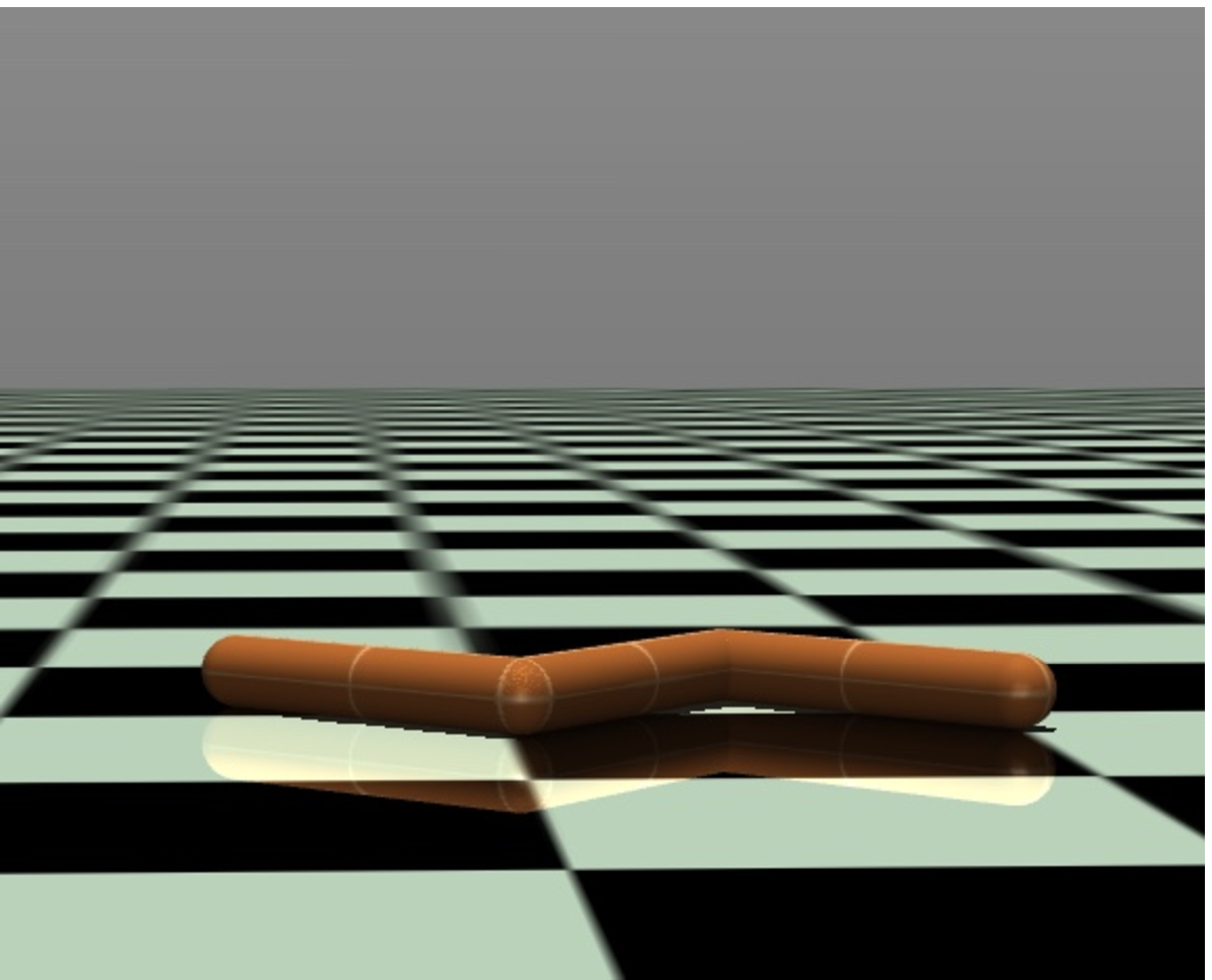}
    \centering
    \subcaption{Swimmer-v2}
    \label{fig:Swimmer-v2_image}
\endminipage\hfill
\centering
\minipage[t]{0.315\linewidth}%
    \includegraphics[height=2cm, width=1\linewidth]{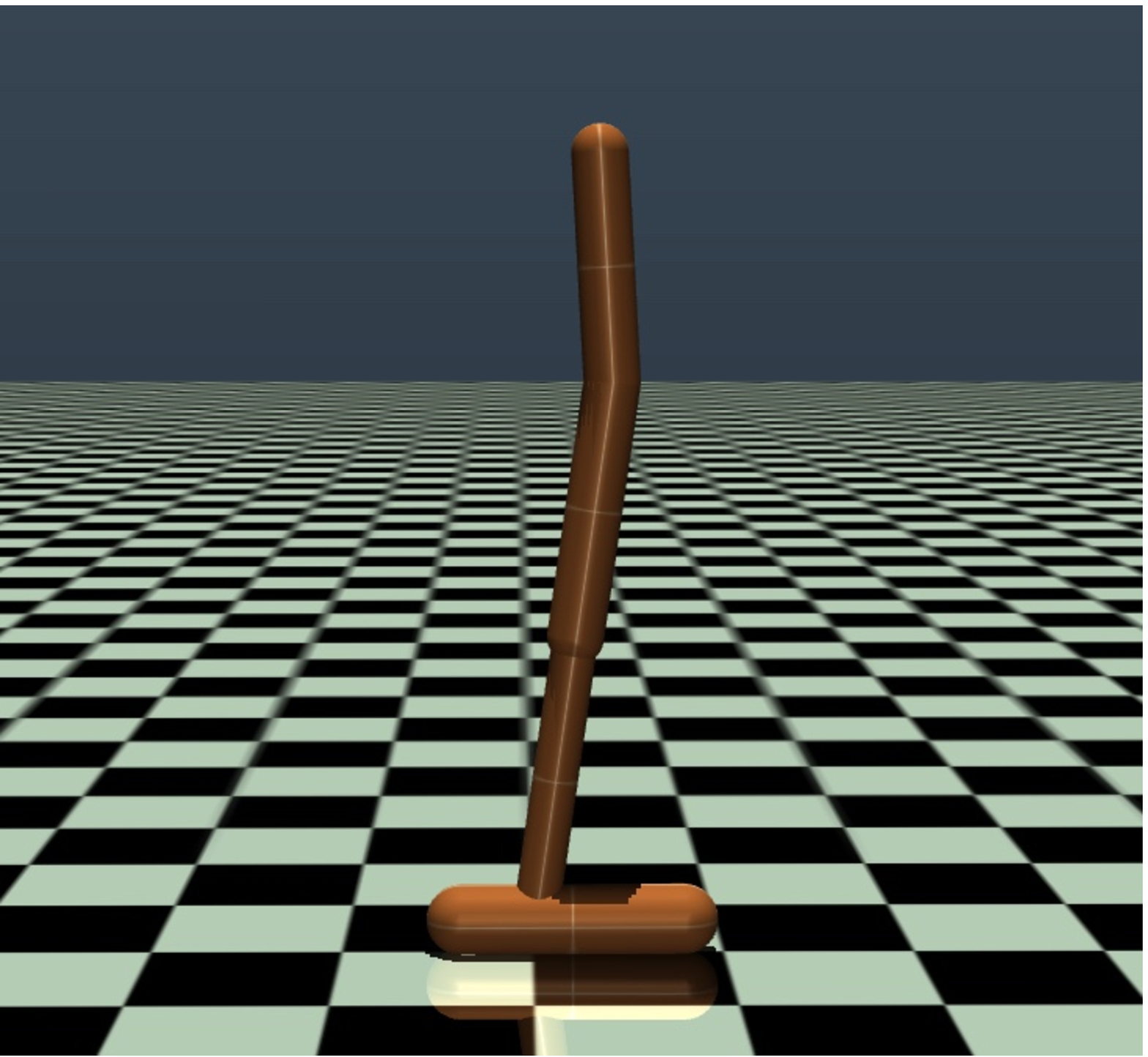}
    \centering
    \subcaption{Hopper-v2}
    \label{fig:Hopper-v2_image}
\endminipage\hfill

\minipage[t]{0.315\linewidth}
    \includegraphics[height=2.1cm, width=1\textwidth]{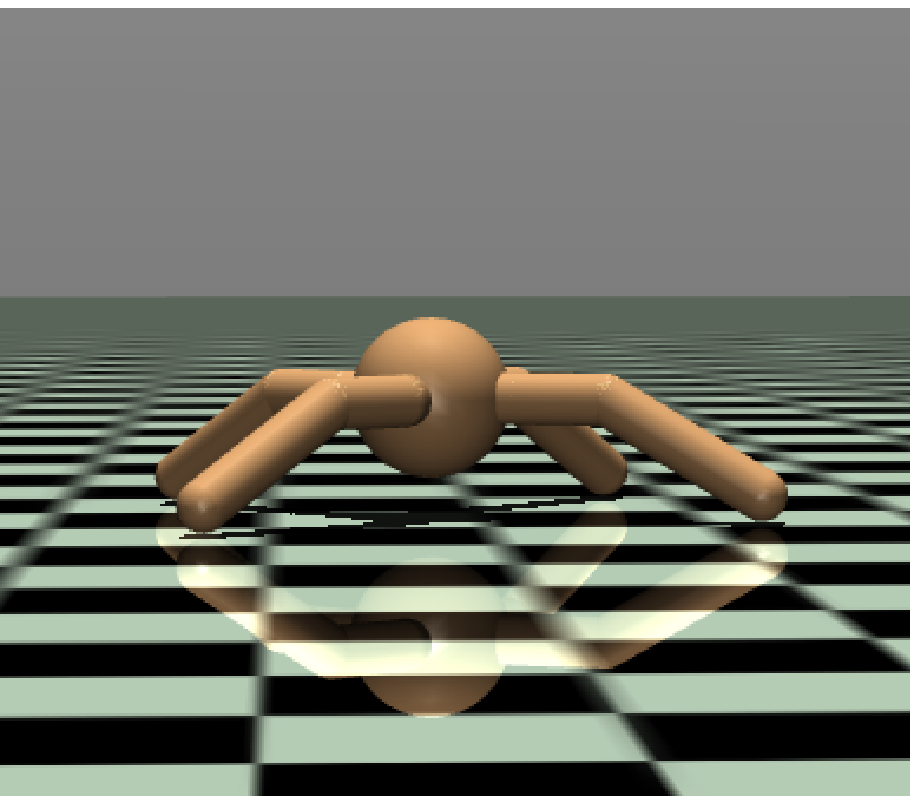}
    \centering
    \subcaption{Ant-v2}
    \label{fig:Ant-v2_image}
\endminipage\hfill
\centering
\minipage[t]{0.33\linewidth}
    \includegraphics[height=2.1cm, width=1\linewidth]{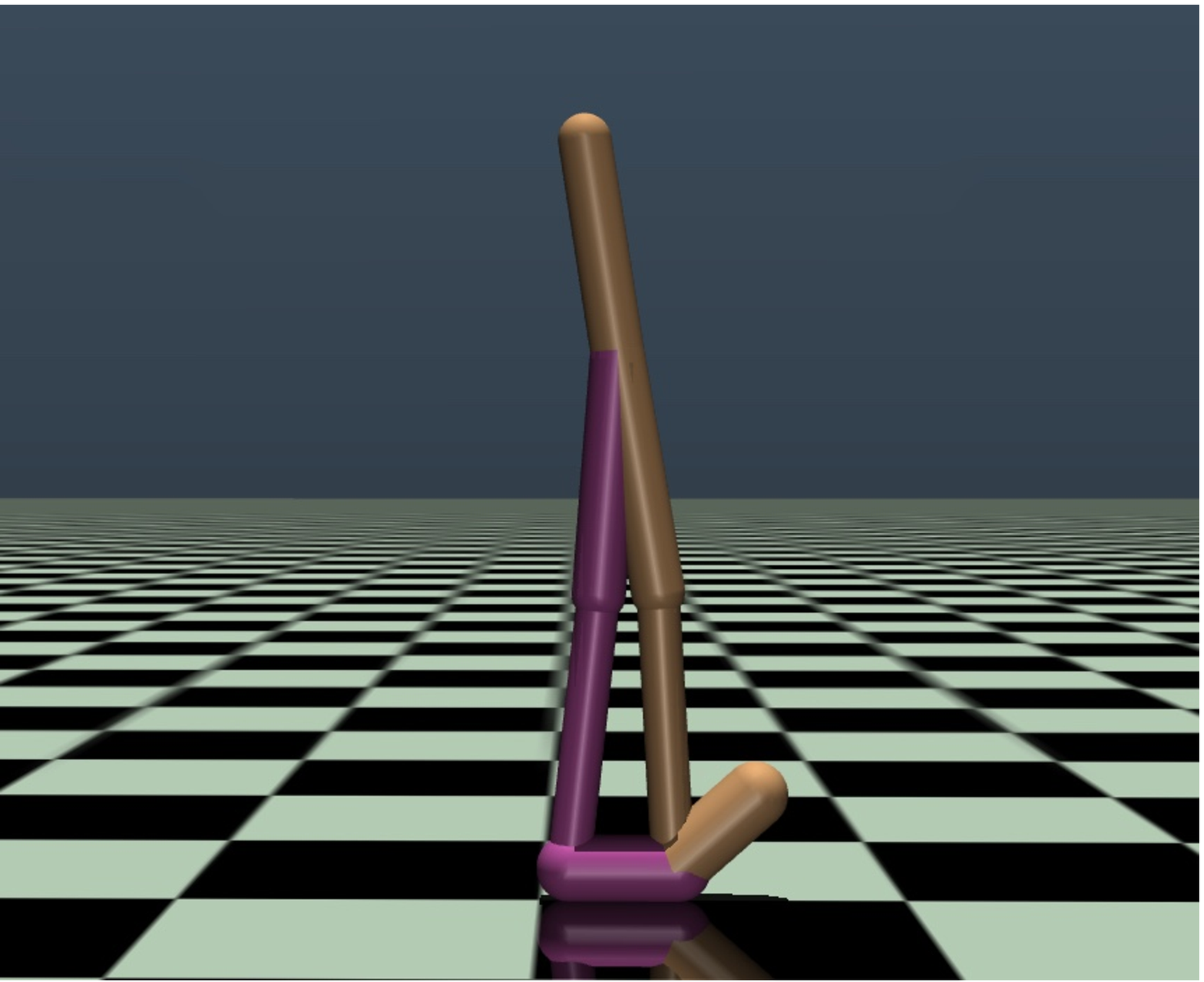}
    \centering
    \subcaption{Walker2d-v2}
    \label{fig:Walker2d-v2_image}
\endminipage\hfill
\centering
\minipage[t]{0.33\linewidth}%
    \includegraphics[height=2.1cm, width=1\linewidth]{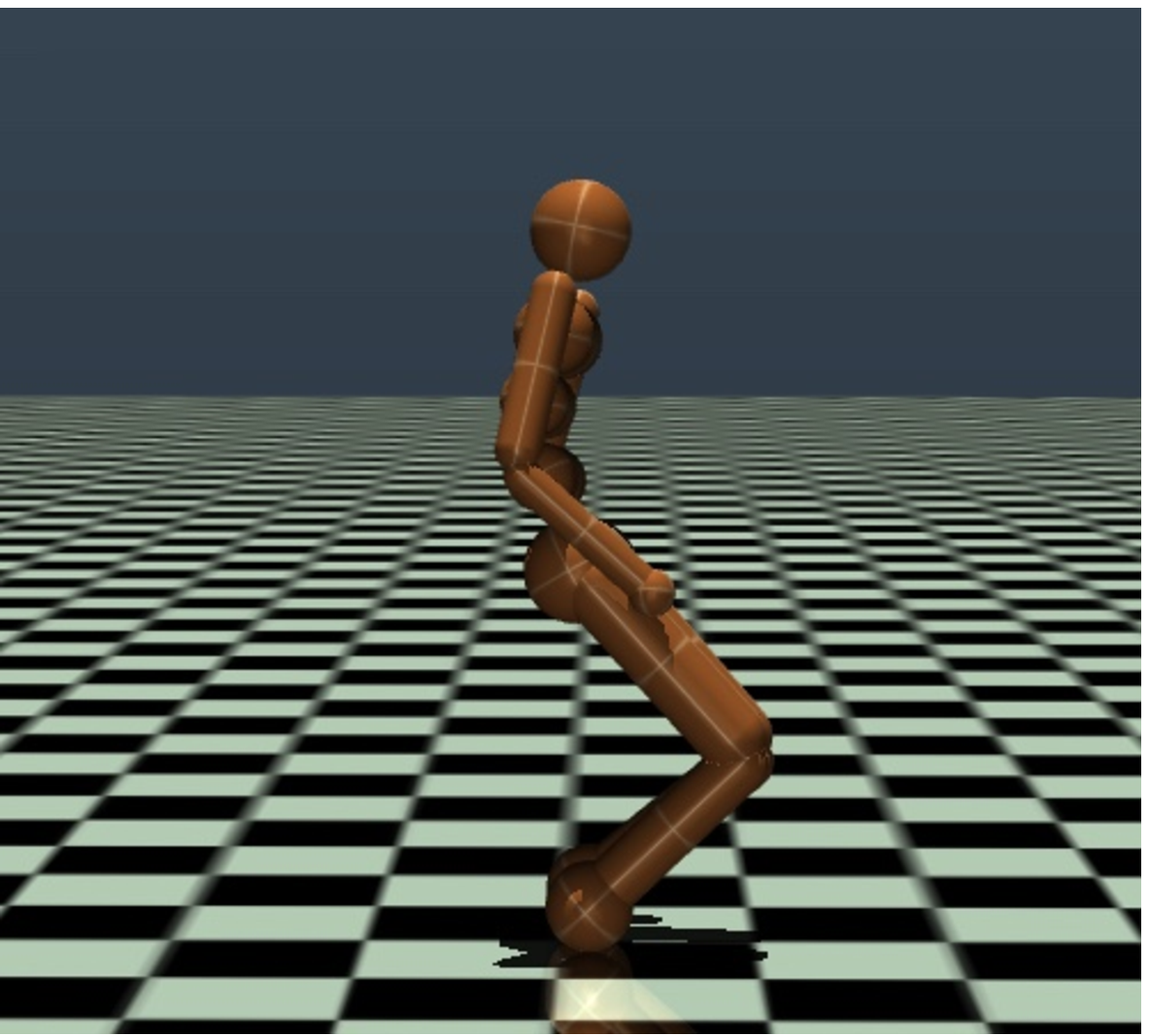}
    \centering
    \subcaption{Humanoid-v2}
    \label{fig:Humanoid-v2_image}
\endminipage\hfill
\caption{Different MuJoCo tasks from the OpenAI Gym.}
\label{fig:mujoco_tasks}
\end{figure}

\begin{figure}[!htb]
    \minipage[t]{0.1523\textwidth}
    \includegraphics[height=2.25cm, width=\textwidth]{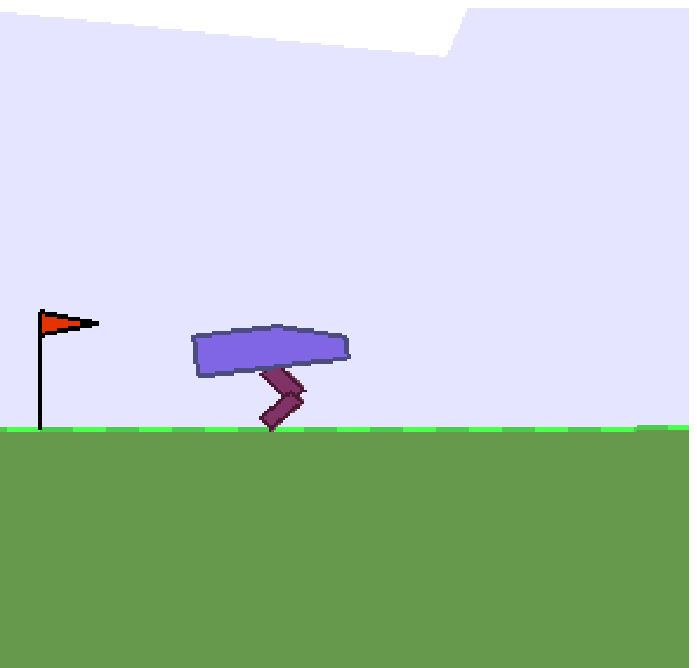}
    \centering
    \label{fig:bipedalwalker_0.5x}
\endminipage\hfill
\centering
\minipage[t]{0.153\textwidth}
    \includegraphics[height=2.25cm, width=\linewidth]{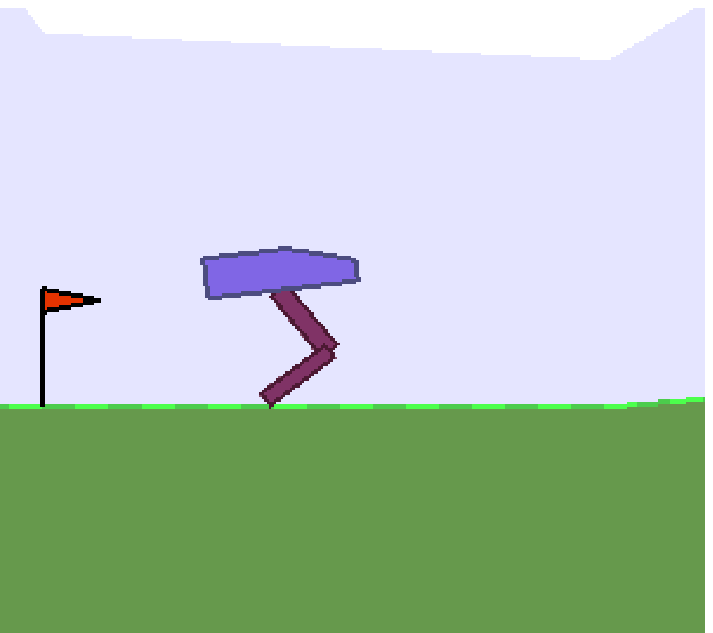}
    \centering
    \label{fig:bipedalwalker_1.0x}
\endminipage\hfill
\centering
\minipage[t]{0.153\textwidth}%
    \includegraphics[height=2.25cm, width=\linewidth]{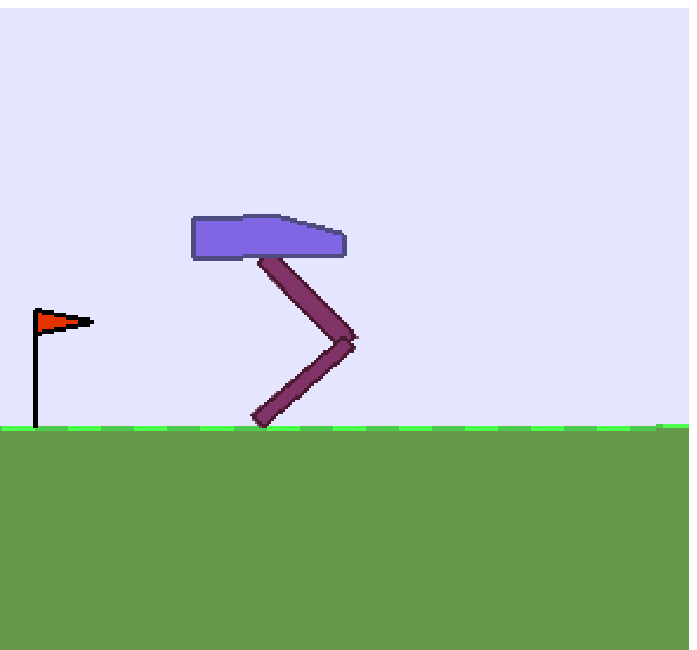}
    \centering
    \label{fig:bipedalwalker_1.5x}
\endminipage\hfill
\caption{Bipedal Walkers with varying leg lengths. From left to right: 50\%, 100\% and 150\% of the default leg length.}
\label{fig:bipedalwalker_simulation}
\end{figure}

In this section, we present a set of experiments on continuous control tasks from the OpenAI Gym \cite{BrockmanCPSSTZ16} (see Figures \ref{fig:mujoco_tasks} and \ref{fig:bipedalwalker_simulation}) to showcase the efficiency of the NuEMT algorithm. 

\subsection{Experimental Configuration}
In our experiments, we compare against the OpenAI-ES \cite{SalimansHCS17} and the recently proposed PEL framework \cite{FuksAHL19}. Our implementation of the latter uses the OpenAI-ES as the base optimizer and is referred to as PEL for the rest of the section. The comparison with the OpenAI-ES allows us to investigate the sample efficiency of our multitask algorithm alongside its single-task counterpart. Similarly, the comparison between sequential transfer (PEL) and multitask transfer (NuEMT) helps us to understand the differences in performance between the two approaches across a variety of environments. This is especially important since we are interested in observing how the limitations of sequential transfers can be averted by multitasking. Recall, the notion of sequential transfer may degrade or stagnate performance if poor or overspecialized solutions are propagated from the simpler to the harder tasks.

The experimental setups are configured as follows. The episode length of the main target task in NuEMT is equal to that of the final task in the PEL baseline; this is equal to the episode length of the single-task in the OpenAI-ES. The number of tasks in NuEMT and PEL are kept the same, for fairness of comparison. For instance, let us assume that there are 3 tasks in NuEMT and PEL, and the full episode length is 1200 timesteps. In the case of NuEMT, the first and second auxiliary tasks will have episode lengths of 400 and 800 timesteps, respectively, while the target task will have maximal episode length of 1200 timesteps. Similarly, for PEL, the episode scheduler is configured as follows. The first task will have an episode length of 400 timesteps, followed by 800 timesteps for the second task, and 1200 timesteps for the last task. In contrast, the single-task OpenAI-ES will have a constant episode length of 1200 timesteps for evaluating all policy parameters generated during its evolutionary run. 

A single run of an algorithm ends when the total number of agent-environment interaction timesteps performed (summed across all tasks in PEL and NuEMT) exceeds a predefined termination condition. This termination condition also determines manual settings of the time scheduler of PEL, i.e., the number of agent-environment interactions performed in each task before moving on to the next. In our implementation, we take the total timesteps for each task in PEL to be uniform (obtained by dividing the termination condition by $K$). Note, we use total timesteps instead of the actual wall-clock time used in \cite{FuksAHL19}.

The total population size $N_{total}$ for all algorithms is the same. In NuEMT, a minimum population size of $N_{total}/K$ is imposed for the main target task along with the adaptive resource allocation strategy, to prevent its population from collapsing---as an exceedingly small population may lead to brittle performance with high variance. For PEL, the population size remains the same for all tasks. The compared algorithms train policies with identical architectures, namely, multilayer perceptrons with 2 hidden layers of 64 nodes and tanh activation functions.

For our implementation, we apply the same parallelization approach in \cite{ManiaGR18} using the Python library Ray \cite{MoritzNWTLLEYPJ18}. All experiments are performed on a single machine with 12-core/24-thread CPU. Each worker holds unique random seeds for sampling noise in the shared noise table as well as initialising the OpenAI Gym environment.

\subsection{Results on Advanced Physics Simulation MuJoCo Tasks}
Here, we compare NuEMT against the baseline algorithms on a variety of MuJoCo tasks. We selected 6 of the popular simulations commonly used in the RL literature, as depicted in Figure \ref{fig:mujoco_tasks}. For the Humanoid-v2 tasks, we find that the survival bonus from the reward function encourages policies that make the MuJoCo models stand at the same spot until maximum episode length is reached \cite{ManiaGR18}, resulting in getting stuck in a local optima of the policy space. To resolve this issue, we minus off the survival bonus (score of 5) from the reward function at each timestep during training. Table \ref{table:mujoco} presents the details of our experiments for each MuJoCo tasks. Our experiment also conducted each simulation for a total of 20 independent trials. In every trial, different random seeds are assigned to each worker and the Gym environment.

\begin{figure}[!htb] 
\centering
\includegraphics[width=0.49\textwidth]{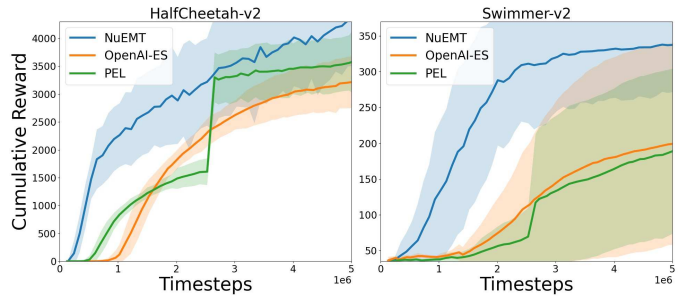}
\caption{Comparison of averaged convergence trends for HalfCheetah-v2 (left) and Swimmer-v2 (right) within a budget of 5 million timesteps of environment interaction.}
\label{fig:mujoco_convergence_trends}
\end{figure}

The results shown in Table \ref{table:mujoco_results} are the mean and standard deviation of the total rewards achieved by all the algorithms at different timesteps. The convergence trends of each algorithm are also shown in \Cref{fig:mujoco_convergence_trends}. We see from Table \ref{table:mujoco_results} that NuEMT outperforms the comparative algorithms at different timesteps for most of the control tasks. Comparing NuEMT with OpenAI-ES, the former is found to offer significant speedup. Note, the main difference between the two algorithms is the transfer of evolved skills in NuEMT. Moreover, the multitask strategy of NuEMT enables accelerated convergence on 5 out of the 6 Mujoco tasks compared to the sequential PEL. The convergence plots in \Cref{fig:mujoco_convergence_trends} reveal a similar story. This is especially clear when we observe convergence behaviours at the initial stages of evolution, where the proposed algorithm rapidly attains higher rewards (on the main task) than the baseline algorithms. Note, on Swimmer-v2, PEL fails to outperform the OpenAI-ES. In contrast, the convergence trends of NuEMT consistently provides strong evidence of its ability to achieve lean evolutionary RL (measured in terms of the amount of agent-environment interaction data needed).

\begin{table*}[!htb]
\caption{Details of the MuJoCo tasks}
\centering
\scalebox{1.0}{
{\renewcommand{\arraystretch}{1.2}
\setlength\tabcolsep{3.3pt}
 \begin{tabular}{c|c|c|c|c|c|c}
 \toprule
 \textbf{Simulation} & \textbf{Algorithm} & \textbf{No. of Tasks} & \textbf{Episode Length} & \textbf{\begin{tabular}[c]{@{}c@{}}Termination Condition\\(No. of Timesteps)\end{tabular}} &
 \textbf{Population Size} &
 \textbf{Learning Parameters}\\ [0.5ex]
 \midrule
 \multirow{3}{*}{HalfCheetah-v2} & NuEMT & \multirow{2}{*}{2} & \multirow{2}{*}{$T_1$: 500, $T_2$: 1000} & \multirow{3}{*}{5,000,000} & \multirow{3}{*}{64} & $\alpha$ = 0.05, $\beta$ = 0.05\\ \cline{2-2}\cline{7-7}
 & PEL & \multirow{2}{*}{} & \multirow{2}{*}{} & \multirow{3}{*}{} & \multirow{3}{*}{} & \multirow{2}{*}{$\alpha$ = 0.05} \\ \cline{2-4} 
 & OpenAI-ES & 1 & 1000 & \multirow{3}{*}{} & \multirow{3}{*}{} & \multirow{2}{*}{} \\ 
 \midrule
 \multirow{3}{*}{Swimmer-v2} & NuEMT & \multirow{2}{*}{2} & \multirow{2}{*}{$T_1$: 500, $T_2$: 1000} & \multirow{3}{*}{5,000,000} & \multirow{3}{*}{64} & $\alpha$ = 0.05, $\beta$ = 0.05\\ \cline{2-2}\cline{7-7}
 & PEL & \multirow{2}{*}{} & \multirow{2}{*}{} & \multirow{3}{*}{} & \multirow{3}{*}{} & \multirow{2}{*}{$\alpha$ = 0.05} \\ \cline{2-4} 
 & OpenAI-ES & 1 & 1000 & \multirow{3}{*}{} & \multirow{3}{*}{} & \multirow{2}{*}{} \\ 
 \midrule
 \multirow{3}{*}{Hopper-v2} & NuEMT & \multirow{2}{*}{2} & \multirow{2}{*}{$T_1$: 500, $T_2$: 1000} & \multirow{3}{*}{16,000,000} & \multirow{3}{*}{64} & $\alpha$ = 0.05, $\beta$ = 0.05\\ \cline{2-2}\cline{7-7}
 & PEL & \multirow{2}{*}{} & \multirow{2}{*}{} & \multirow{3}{*}{} & \multirow{3}{*}{} & \multirow{2}{*}{$\alpha$ = 0.05} \\ \cline{2-4} 
 & OpenAI-ES & 1 & 1000 & \multirow{3}{*}{} & \multirow{3}{*}{} & \multirow{2}{*}{} \\ 
 \midrule
 \multirow{3}{*}{Ant-v2} & NuEMT & \multirow{2}{*}{2} & \multirow{2}{*}{$T_1$: 500, $T_2$: 1000} & \multirow{3}{*}{9,000,000} & \multirow{3}{*}{128} & $\alpha$ = 0.05, $\beta$ = 0.05\\ \cline{2-2}\cline{7-7}
 & PEL & \multirow{2}{*}{} & \multirow{2}{*}{} & \multirow{3}{*}{} & \multirow{3}{*}{} & \multirow{2}{*}{$\alpha$ = 0.05} \\ \cline{2-4} 
 & OpenAI-ES & 1 & 1000 & \multirow{3}{*}{} & \multirow{3}{*}{} & \multirow{2}{*}{} \\ 
 \midrule
 \multirow{3}{*}{Walker2d-v2} & NuEMT & \multirow{2}{*}{2} & \multirow{2}{*}{$T_1$: 500, $T_2$: 1000} & \multirow{3}{*}{50,000,000} & \multirow{3}{*}{96} & $\alpha$ = 0.05, $\beta$ = 0.05\\ \cline{2-2}\cline{7-7}
 & PEL & \multirow{2}{*}{} & \multirow{2}{*}{} & \multirow{3}{*}{} & \multirow{3}{*}{} & \multirow{2}{*}{$\alpha$ = 0.05} \\ \cline{2-4} 
 & OpenAI-ES & 1 & 1000 & \multirow{3}{*}{} & \multirow{3}{*}{} & \multirow{2}{*}{} \\ 
 \midrule
 \multirow{3}{*}{Humanoid-v2} & NuEMT & \multirow{2}{*}{4} & \multirow{2}{*}{$T_1$: 250, $T_2$: 500, $T_3$: 750, $T_4$: 1000} & \multirow{3}{*}{40,000,000} & \multirow{3}{*}{480} & $\alpha$ = 0.1, $\beta$ = 0.05\\ \cline{2-2}\cline{7-7}
 & PEL & \multirow{2}{*}{} & \multirow{2}{*}{} & \multirow{3}{*}{} & \multirow{3}{*}{} & \multirow{2}{*}{$\alpha$ = 0.1} \\ \cline{2-4} 
 & OpenAI-ES & 1 & 1000 & \multirow{3}{*}{} & \multirow{3}{*}{} & \multirow{2}{*}{} \\ 
 \bottomrule
\end{tabular}}
}

\label{table:mujoco}
\end{table*}

\begin{table*}[!htb]
\caption{Comparison of the mean performance and standard deviation across Mujoco tasks. Results in bold indicate best mean performance. For comparisons of the OpenAI-ES against standard policy gradient methods, readers are referred to \cite{SalimansHCS17}.}
\centering
\scalebox{1}{
{\renewcommand{\arraystretch}{1.2}
\setlength\tabcolsep{3.5pt}
 \begin{tabular}{c|c|c|c|c}
 \toprule
 \multirow{2}{*}{\textbf{Simulation}} &  \multirow{2}{*}{\textbf{No. of Timesteps}} & \textbf{NuEMT} & \textbf{OpenAI-ES} & \textbf{PEL} \\ \cline{3-5}
 \multirow{2}{*}{} & \multirow{2}{*}{} & \textbf{Mean $\pm$ Std} & \textbf{Mean $\pm$ Std} & \textbf{Mean $\pm$ Std} \\
 \midrule
  \multirow{2}{*}{HalfCheetah-v2} & 2,500,000 & \textbf{3241.65 $\pm$ 780.85} & 2278.48 $\pm$ 237.24 & 1610.29 $\pm$ 243.84 \\ \cline{2-5}
  \multirow{2}{*}{} & 5,000,000 & \textbf{4291.73 $\pm$ 822.79} & 3217.82 $\pm$ 469.71 & 3572.97 $\pm$ 541.49 \\
 \midrule
  \multirow{2}{*}{Swimmer-v2} & 2,500,000 & \textbf{334.82 $\pm$ 33.45} & 109.29 $\pm$ 83.05 & 67.79 $\pm$ 38.22 \\ \cline{2-5}
  \multirow{2}{*}{} & 5,000,000 & \textbf{359.54 $\pm$ 13.50} & 199.07 $\pm$ 141.60 & 188.81 $\pm$ 115.43 \\
 \midrule
  \multirow{2}{*}{Hopper-v2} & 8,000,000 & \textbf{2157.65 $\pm$ 1015.59} & 1178.39 $\pm$ 34.01 & 1265.63 $\pm$ 332.43 \\ \cline{2-5}
  \multirow{2}{*}{} & 16,000,000 & 3003.78 $\pm$ 895.22 & 1304.84 $\pm$ 478.82 & \textbf{3223.41 $\pm$ 543.70} \\
 \midrule
  \multirow{2}{*}{Ant-v2} & 4,500,000 & \textbf{2057.33 $\pm$ 118.22} & 1127.06 $\pm$ 57.54 & 833.76 $\pm$ 56.51 \\ \cline{2-5}
  \multirow{2}{*}{} & 9,000,000 & \textbf{2513.13 $\pm$ 186.28} & 1615.80$\pm$ 81.69 & 2131.40 $\pm$ 213.71 \\
 \midrule
  \multirow{2}{*}{Walker2d-v2} & 25,000,000 & \textbf{4176.12 $\pm$ 415.54} & 1604.66 $\pm$ 664.29 & 1644.01 $\pm$ 363.12 \\ \cline{2-5}
  \multirow{2}{*}{} & 50,000,000 & \textbf{4746.42 $\pm$ 477.50} & 3871.28 $\pm$ 669.72 & 4739.50 $\pm$ 955.66 \\
  \midrule
  \multirow{4}{*}{Humanoid-v2*} & 10,000,000 & \textbf{169.56 $\pm$ 62.46} & 44.89 $\pm$ 1.55 & 44.89 $\pm$ 1.55 \\ \cline{2-5}
  \multirow{4}{*}{} & 20,000,000 & \textbf{241.08 $\pm$ 91.81} & 59.99 $\pm$ 8.37 & 59.99 $\pm$ 8.37 \\ \cline{2-5}
  \multirow{4}{*}{} & 30,000,000 & \textbf{297.27 $\pm$ 107.08} & 94.20 $\pm$ 20.86 & 94.20 $\pm$ 20.86 \\ \cline{2-5}
  \multirow{4}{*}{} & 40,000,000 & \textbf{345.29 $\pm$ 108.80} & 136.92 $\pm$ 53.06 & 136.92 $\pm$ 53.06 \\
 \bottomrule

\end{tabular}}
}
\vspace{0.2cm}
\begin{tablenotes}
    \footnotesize
    \item *Survival bonus is removed at every timestep. OpenAI-ES and PEL perform identically since none of the OpenAI-ES's agents survive beyond the configured episode length for PEL throughout the training.
\end{tablenotes}
\label{table:mujoco_results}
\end{table*}

\subsection{Results on Box2d Simulators}

Here, we selected BipedalWalker as a representative in the Box2D simulator from OpenAI Gym. In the BipedalWalker experiment, the robot gets up to 300 plus points when it reaches the far end and 100 points are deducted if the robot falls. We have adjusted the length of the legs (depicted in Figure \ref{fig:bipedalwalker_simulation}) such that each continuous control task will have different stability issues to overcome while traversing the terrain. Table \ref{table:box2d} shows the number of tasks, episode length, number of timesteps (termination condition) and other implementation details for our experiments. For each simulation, a total of 20 independent trials are conducted. Similar to the MuJoCo experiments, unique random seeds are assigned to each worker and the Gym environment in every trial.

Table \ref{table:box2d_results} shows the mean and standard deviations of the total rewards at different timesteps. As seen from the results, NuEMT outperforms the comparative algorithms at most timesteps in all three simulations. Comparing NuEMT with the base optimizer, OpenAI-ES, the former is found to be significantly more data-efficient. This improvement is a consequence of inter-task skills transfer, which is the fundamental algorithmic distinction between NuEMT and OpenAI-ES. The superior sample efficiency is achieved by tapping on useful information from simpler tasks (of shorter agent-environment interaction episodes) to quickly achieve better performance on the longer and harder task at hand. 

While the PEL baseline performs slightly better than NuEMT on the 0.5x leg length BipedalWalker-v3 simulation, it struggles to perform consistently in the other two simulations that pose increasing difficulty in maintaining balance due to longer leg lengths. In contrast, the NuEMT methodology maintains sample efficient performance in all three experiments, vastly outperforming competitors on 1.0x and 1.5x leg length. 

\begin{table*}[ht]
\caption{Details of the Box2d Simulators}
\centering
\scalebox{1.0}{
{\renewcommand{\arraystretch}{1.2}
\setlength\tabcolsep{3.3pt}
 \begin{tabular}{c|c|c|c|c|c|c}
 \toprule
 \textbf{Simulation} & \textbf{Algorithm} & \textbf{No. of Tasks} & \textbf{Episode Length} & \textbf{\begin{tabular}[c]{@{}c@{}}Termination Condition\\(No. of Timesteps)\end{tabular}} &
 \textbf{Population Size} &
 \textbf{Learning Parameters}\\ [0.5ex]
 \midrule
 \multirow{3}{*}{\begin{tabular}[c]{@{}c@{}}BipedalWalker-v3\\(0.5x Leg Length**)\end{tabular}} & NuEMT & \multirow{2}{*}{4} & \multirow{2}{*}{$T_1$: 400, $T_2$: 800, $T_3$: 1200, $T_4$: 1600} & \multirow{3}{*}{70,000,000} & \multirow{3}{*}{128} & $\alpha$ = 0.05, $\beta$ = 0.05\\ \cline{2-2}\cline{7-7}
 & PEL & \multirow{2}{*}{} & \multirow{2}{*}{} & \multirow{3}{*}{} & \multirow{3}{*}{} & \multirow{2}{*}{$\alpha$ = 0.05} \\ \cline{2-4} 
 & OpenAI-ES & 1 & 1600 & \multirow{3}{*}{} & \multirow{3}{*}{} & \multirow{2}{*}{} \\ 
 \midrule
 \multirow{3}{*}{\begin{tabular}[c]{@{}c@{}}BipedalWalker-v3\\(1.0x Leg Length**)\end{tabular}} & NuEMT & \multirow{2}{*}{4} & \multirow{2}{*}{$T_1$: 400, $T_2$: 800, $T_3$: 1200, $T_4$: 1600} & \multirow{3}{*}{80,000,000} & \multirow{3}{*}{128} & $\alpha$ = 0.05, $\beta$ = 0.05\\ \cline{2-2}\cline{7-7}
 & PEL & \multirow{2}{*}{} & \multirow{2}{*}{} & \multirow{3}{*}{} & \multirow{3}{*}{} & \multirow{2}{*}{$\alpha$ = 0.05} \\ \cline{2-4} 
 & OpenAI-ES & 1 & 1600 & \multirow{3}{*}{} & \multirow{3}{*}{} & \multirow{2}{*}{} \\ 
 \midrule
 \multirow{3}{*}{\begin{tabular}[c]{@{}c@{}}BipedalWalker-v3\\(1.5x Leg Length**)\end{tabular}} & NuEMT & \multirow{2}{*}{4} & \multirow{2}{*}{$T_1$: 400, $T_2$: 800, $T_3$: 1200, $T_4$: 1600} & \multirow{3}{*}{90,000,000} & \multirow{3}{*}{128} & $\alpha$ = 0.05, $\beta$ = 0.05\\ \cline{2-2}\cline{7-7}
 & PEL & \multirow{2}{*}{} & \multirow{2}{*}{} & \multirow{3}{*}{} & \multirow{3}{*}{} & \multirow{2}{*}{$\alpha$ = 0.05} \\ \cline{2-4} 
 & OpenAI-ES & 1 & 1600 & \multirow{3}{*}{} & \multirow{3}{*}{} & \multirow{2}{*}{} \\ 
 \bottomrule
\end{tabular}}
}
\vspace{0.2cm}
\begin{tablenotes}
    \footnotesize
    \item **Bipedal walkers with varying leg lengths are shown in Figure \ref{fig:bipedalwalker_simulation}.
\end{tablenotes}
\label{table:box2d}
\end{table*}

\begin{table*}[ht]
\caption{Comparison of the mean performance and standard deviation across Box2d tasks and comparative algorithms. Results in bold indicate best mean performance.}
\centering
\scalebox{1}{
{\renewcommand{\arraystretch}{1.2}
\setlength\tabcolsep{3.5pt}
 \begin{tabular}{c|c|c|c|c}
 \toprule
 \multirow{2}{*}{\textbf{Simulation}} &  \multirow{2}{*}{\textbf{No. of Timesteps}} & \textbf{NuEMT} & \textbf{OpenAI-ES} & \textbf{PEL} \\ \cline{3-5}
 \multirow{2}{*}{} & \multirow{2}{*}{} & \textbf{Mean $\pm$ Std} & \textbf{Mean $\pm$ Std} & \textbf{Mean $\pm$ Std} \\
 \midrule
  \multirow{4}{*}{\begin{tabular}[c]{@{}c@{}} BipedalWalker-v3\\(0.5x Leg Length)\end{tabular}} & 17,500,000 & \textbf{282.92 $\pm$ 38.13} & 58.81  $\pm$ 89.33 & 67.27 $\pm$ 23.20 \\ \cline{2-5}
  \multirow{4}{*}{} & 35,000,000 & \textbf{295.55 $\pm$ 19.09} & 100.70 $\pm$ 120.17 & 206.39 $\pm$ 51.93  \\ \cline{2-5}
  \multirow{4}{*}{} & 52,500,000 & 296.13 $\pm$ 14.34 & 123.03 $\pm$ 126.98 & \textbf{302.68 $\pm$ 40.02}  \\ \cline{2-5}
  \multirow{4}{*}{} & 70,000,000 & 295.11 $\pm$ 12.91 & 181.70 $\pm$ 118.19 & \textbf{331.06 $\pm$ 13.60}  \\
 \midrule
  \multirow{4}{*}{\begin{tabular}[c]{@{}c@{}} BipedalWalker-v3\\(1.0x Leg Length)\end{tabular}} & 20,000,000 & \textbf{233.17 $\pm$ 74.27} & 5.18 $\pm$ 3.37 & 8.61 $\pm$ 0.66 \\ \cline{2-5}
  \multirow{4}{*}{} & 40,000,000 & \textbf{262.94 $\pm$ 73.06} & 6.47 $\pm$ 2.45 & 11.31 $\pm$ 8.91 \\ \cline{2-5}
  \multirow{4}{*}{} & 60,000,000 & \textbf{290.61 $\pm$ 19.22} & 7.85 $\pm$ 3.24 & 13.92 $\pm$ 12.29  \\ \cline{2-5}
  \multirow{4}{*}{} & 80,000,000 & \textbf{297.38 $\pm$ 15.47} & 8.15 $\pm$ 3.64 & 44.64 $\pm$ 80.10 \\
 \midrule
  \multirow{3}{*}{\begin{tabular}[c]{@{}c@{}}BipedalWalker-v3\\(1.5x Leg Length)\end{tabular}} & 22,500,000 & \textbf{27.72 $\pm$ 66.12} & 10.09 $\pm$ 4.67 & 13.34 $\pm$ 0.17  \\ \cline{2-5}
  \multirow{3}{*}{} & 45,000,000 & \textbf{80.43 $\pm$ 115.70} & 11.35 $\pm$ 4.53 & 13.03 $\pm$ 0.36 \\ \cline{2-5}
  \multirow{3}{*}{} & 67,500,000 & \textbf{133.33 $\pm$ 128.44} & 12.46 $\pm$ 0.415 & 12.83 $\pm$ 0.53 \\ \cline{2-5}
  \multirow{3}{*}{} & 90,000,000 & \textbf{152.28 $\pm$ 127.35} & 12.79 $\pm$ 0.52 & 12.68 $\pm$ 0.57 \\
 \bottomrule
\end{tabular}}
}

\label{table:box2d_results}
\end{table*}

\begin{figure}[!htb] 
\centering
\includegraphics[width=0.49\textwidth]{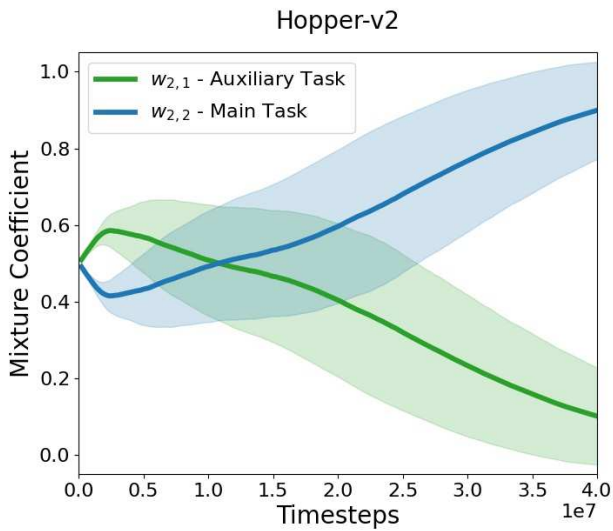}
\caption{Historical update trends of the mixture coefficients for the main and auxiliary tasks in the Hopper-v2 environment. The shaded area represents one standard deviation on either side of the mean. The convergence trends align well with the conclusion drawn at the end of Section IV-D.}
\label{fig:historical_update_mixture_co}
\end{figure}

\subsection{Analysing the Behavior of NuEMT}

\subsubsection{Update History of the Mixture Coefficients} During the initial stages of evolution, if task $T_i$ transfers useful skills to the main task $T_K$, we expect the mixture coefficient $w_{K,i}$ to increase, allocating greater computational resources to $T_i$ to quickly discover better solutions at lower cost. However, at later stages, after the target probabilistic model $p_K(\theta)$ has arrived at a promising region of the search space, the mixture coefficient value $w_{K, K}$ is expected to increase (whereas $w_{K,i}$ is expected to drop), thus increasing computational effort on refinement of the target search. To verify this intuition, we perform 20 independent trials in the Hopper-v2 environment (for an extended period of 40 million timesteps) to investigate the temporal latent relationship between the main task and the auxiliary task by observing the historical updates of the learnt mixture coefficients. Averaged convergence plots are shown in Figure \ref{fig:historical_update_mixture_co}, and are found to not only align qualitatively with our intuition, but also with the theoretical conclusion drawn at the end of Section IV-D.

\subsubsection{Effectiveness of Mixture Coefficient Learning}
\begin{figure}[!htb]
\begin{center}
\subfloat[HalfCheetah-v2]{%
   \includegraphics[height=6cm, width=0.44\textwidth]{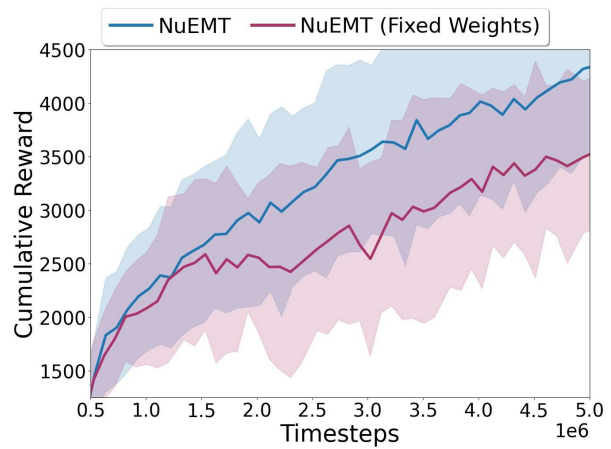}%
  \centering
}

\subfloat[Hopper-v2]{%
  \includegraphics[height=6cm, width=0.44\textwidth]{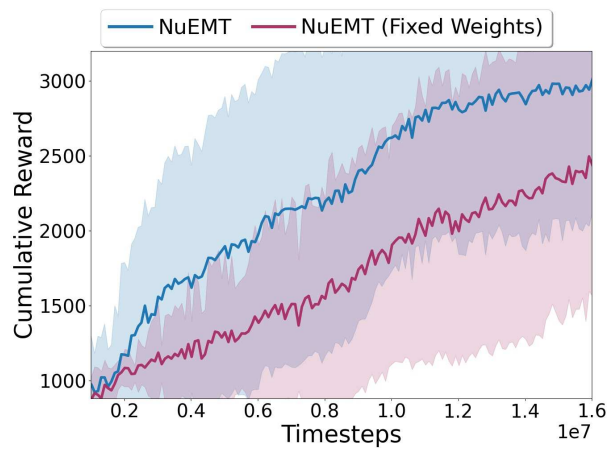}%
  \centering
}
\caption{Performance of NuEMT with learnt mixture coefficients versus NuEMT with fixed mixture coefficients. Shaded area represents one standard deviation on either side of the mean.}
\label{fig:effects_of_mixture_learning}
\end{center}
\end{figure}
It is worth understanding how our algorithm would perform with and without mixture coefficient learning. To this end, we perform two sets of experiments on HalfCheetah-V2 and Hopper-v2, two of the MuJoCo simulations. One experiment is performed with our mixture coefficient learning and subsequent mixture coefficient update, while the other experiment is performed with a fixed set of mixture coefficients that does not change over iterations. The number of auxiliary tasks is set to 1, as per the configuration in earlier results. The mean performance between the two sets of experiment is shown in Figure \ref{fig:effects_of_mixture_learning}. We see that mixture coefficient learning in Eq. (\ref{eq:weights_update}) is able to uplift the performance of the fixed weight variant by up to 23\% at the end of the training curve. This provides an interesting outlook of how mixture coefficients can directly affect the gradient estimate for $\tilde{\theta}$ (see Eq. (\ref{theta_update})), driving search towards good solutions by \emph{self-adapting} the sampling between different search distribution models.

\subsubsection{Analysing Effects of the Number of Auxiliary Tasks}

\begin{figure}[!htb] 
\centering
\includegraphics[height=5.75cm, width=0.48\textwidth]{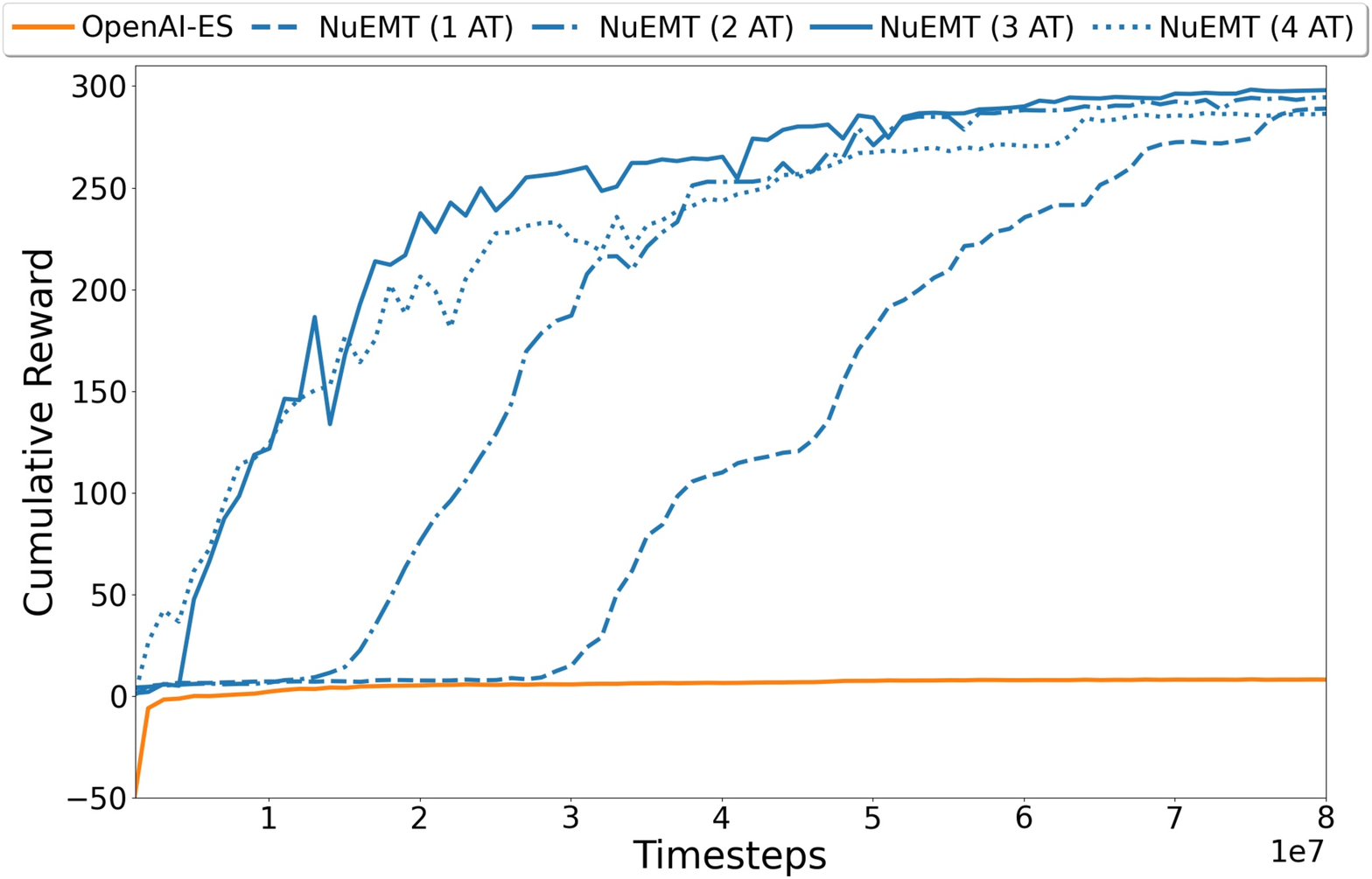}
\caption{Comparison of the mean performance trends achieved using different numbers of auxiliary tasks on the BipedalWalker-v3 simulation. The total population size is 128 candidate solutions. The term `auxiliary tasks' is abbreviated as AT in the legend.}
\label{fig:effects_of_auxiliary_tasks}
\end{figure}

Here, we investigate how the number of auxiliary tasks in NuEMT affects its performance. We ran several experiments with different numbers of auxiliary tasks in the BipedalWalker-v3 simulation. The Bipedalwalker-v3 simulation used has the default leg length. For the experimental settings, each experiment consists of 20 independent trials with 20 different seeds. The population size is set to be 128 candidate solutions. The learning rates $\alpha, \beta$ for parameter $\theta$ and mixture coefficient $w$ are fixed as 0.05 and 0.05, respectively. Our goal is to analyse the effect of increasing the number of tasks given a fixed population size. Averaged results are shown in Figure \ref{fig:effects_of_auxiliary_tasks}.

From Figure \ref{fig:effects_of_auxiliary_tasks}, we see that NuEMT with 3 auxiliary tasks performs the best among the 5 line plots, followed by NuEMT with 4 auxiliary tasks, NuEMT with 2 auxiliary tasks, NuEMT with 1 auxiliary task, and finally NuEMT with no auxiliary task (which reduces to the OpenAI-ES). While having more auxiliary tasks could provide more information to the target, the total population size must also grow to allow the search on shorter episode lengths to generate useful transferrable skills. Hence, with a fixed population size of 128 solutions, NuEMT with 3 auxiliary tasks managed to outperform NuEMT with 4 auxiliary tasks in the same environment.

\section{Conclusion}
In this paper, we explored the application of evolutionary multitasking as a novel means to achieve data-efficient evolutionary RL. Our proposed \emph{neuroevolutionary multitasking} (NuEMT) algorithm is based on the idea of harnessing useful information (transferrable skills) from auxiliary tasks with shorter episode lengths, to quickly optimize a neural network policy for the target task at hand. The uniqueness of NuEMT lies in utilizing the statistical \emph{importance sampling} technique as the information transfer mechanism within the base optimizer, OpenAI-ES, without having to modify its other search operators. The multitasking \emph{trick} is shown to provide enhanced sample efficiency, attaining higher cumulative rewards with lesser agent-environment interactions. 

In our experiments, a variety of continuous control environments from the OpenAI Gym were considered. The results unveiled significant advantages of multitasking over the single-task OpenAI-ES as well as a sequential transfer-based ES (which made use of the same auxiliary tasks). Multitasking overcomes the threat faced by sequential transfer in those cases where solutions evolved for shorter episodes do not propagate well to longer and harder tasks in the future. Our results thus mark a major step forward in confirming the viability of evolutionary algorithms as simple, scalable and sample efficient alternatives for deep RL. 

For the next step in this line of research, we plan to extend the general idea of leveraging simpler tasks to improve learning on complex problems beyond the realms of RL. Other machine learning sub-fields, such as neural architecture search, may also benefit greatly from the potential to jointly evolve multiple tasks, producing diverse models specialized to different datasets and/or different hardware constraints in a single evolutionary run. 

\bibliographystyle{IEEEtran}
\bibliography{IEEEbib}

\end{document}